\documentclass[10pt,twocolumn,letterpaper]{article}

\usepackage[pagenumbers]{wacv} % To force page numbers, e.g. for an arXiv version

\usepackage{xcolor}
\newcommand{\red}[1]{{\color{red}#1}}

\newcommand{\blue}[1]{\textcolor{blue}{#1}}

\usepackage{graphicx}
\usepackage{booktabs}
\usepackage{amsmath}   % equation, \lVert
\usepackage{bm}        % \bm
\usepackage[table]{xcolor}
\definecolor{ldgray}{RGB}{245,245,245}
\definecolor{ldblue}{RGB}{220,243,255}
\usepackage{tabularx}
\usepackage{multirow}
\usepackage{comment}
\usepackage{subcaption}
\usepackage{float}
\usepackage[accsupp]{axessibility}  % Improves PDF readability for those with disabilities.

\usepackage{xcolor}
\definecolor{wacvblue}{rgb}{0.21,0.49,0.74}
\usepackage[pagebackref,breaklinks,colorlinks,allcolors=wacvblue]{hyperref}

\def\wacvPaperID{542} % *** Enter the WACV Paper ID here
\def\confName{WACV}
\def\confYear{2027}

\title{FDSML: Formula-Driven Supervised Metric Learning\\with Parameter-Aware Triplet Loss}

\author{
Hina Otake$^{1\ast}$ \quad
Shinei Arakawa$^{1\ast}$ \quad
Keitaro Tanaka$^{2\ast}$ \\
Yoshihiro Fukuhara$^{1,3}$ \quad
Hirokatsu Kataoka$^{3,4}$ \quad
Shigeo Morishima$^{2}$ \\[4pt]
$^{1}$Waseda University \quad
$^{2}$Waseda Research Institute for Science and Engineering \\
$^{3}$National Institute of Advanced Industrial Science and Technology \quad
$^{4}$University of Oxford
}

\begin{document}
\AddToShipoutPicture*{%
  \AtPageUpperLeft{%
    \raisebox{-0.45in}{%
      \makebox[\paperwidth]{%
        \parbox[t]{\textwidth}{\scriptsize\centering\color{gray}
          \copyright~2027 IEEE. Personal use of this material is permitted.
          Permission from IEEE must be obtained for all other uses, in any current or future media, 
          including reprinting/republishing this material for advertising or promotional purposes, creating new collective works, for resale or redistribution to servers or lists, or reuse of any copyrighted component of this work in other works.}}}}}
\maketitle

\begingroup
\renewcommand{\thefootnote}{}%
\footnotetext{$^{\ast}$Equal contribution.}%
\endgroup

\begin{abstract}
Formula-Driven Supervised Learning (FDSL) enables vision pre-training without real images, but existing methods cast pre-training as classification on discretized labels, discarding continuity in the generative process.
We propose Formula-Driven Supervised Metric Learning (FDSML), 
which derives relational supervision from image-generation parameters beyond discrete class labels.
FDSML combines FineFractalDB, our refined fractal dataset with higher intra-class variation and more stable structures, with a Parameter-Aware Triplet (PA-Triplet) Loss that assigns positives and negatives according to distances in the underlying Iterated Function System parameter space.
This objective trains embeddings to respect parameter-derived relative similarity orderings.
Experiments show that FDSML improves transferability to image retrieval while remaining competitive on downstream classification and improving robustness under several corruption settings.
On ResNet-50, Recall@1 improves from 65.6\% to 74.0\% on Stanford Online Products and from 56.2\% to 78.4\% on Stanford Cars over FDSL pre-training.
These results suggest that parameter-derived relational supervision provides a stronger synthetic pre-training signal than label-based classification for transfer beyond recognition.
Project page: \url{https://shineiarakawa.github.io/fdsml}.
\end{abstract}
    
\section{Introduction}
\label{sec:intro}

The success of modern computer vision models relies heavily on large-scale, human-annotated datasets such as ImageNet~\cite{deng2009imagenet}.
However, the paradigm of real-world pre-training faces escalating challenges, including prohibitive annotation costs~\cite{lin2014microsoft, krishna2016visualgenomeconnectinglanguage}, privacy concerns~\cite{yang2022study, ren2018learning, asano21pass}, and inherent social biases~\cite{de2019does, birhane2021large, tommasi2017deeper}.
To circumvent these bottlenecks, synthetic image pre-training has emerged as a promising alternative~\cite{richter2016playing, tobin2017domain, tian2024learning}, offering an infinitely scalable and privacy-compliant source of visual data.

Among synthetic approaches, Formula-Driven Supervised Learning (FDSL)~\cite{Kataoka_2020_ACCV, Nakashima_Kataoka_Matsumoto_Iwata_Inoue_Satoh_2022, shinoda2023segrcdb, 9880154, Kataoka_2021_ICCV, Takashima_2023_CVPR, Nakamura_2023_ICCV, Nakamura_2024_ECCV}, exemplified by FractalDB~\cite{Kataoka_2020_ACCV}, has shown potential.
By generating images from mathematical formulas and assigning labels based on the generation parameters, FDSL allows models to learn complex visual patterns without any natural images.
FDSL pre-training has shown competitive or even superior performance on downstream classification tasks~\cite{lecun2002gradient, dosovitskiy2020vit}.

Despite this success, a fundamental question remains: \textbf{does FDSL truly replace real-world pre-training, or merely imitate it?}
While FDSL-trained models can achieve high classification accuracy, 
their learned representations can still differ from those learned on real data.
This limitation becomes more apparent in tasks such as image retrieval, where performance relies on the quality of feature representations rather than classification boundaries.
Furthermore, these models can be sensitive to perturbations such as blur and can even underperform training from scratch in certain target domains.

We argue that this limitation stems from biases inherent in mathematics-driven image generation.
Traditional FDSL tasks models with identifying synthetic categories.
Since synthetic images differ structurally from natural scenes, models tend to overfit to localized, low-level geometric textures.
To address this issue, we propose a paradigm shift in FDSL: from simple recognition to relational comparison, encouraging models to learn \textbf{how to look} at visual structures and their relationships by reasoning about relative distances among samples.

To this end, we propose \textbf{Formula-Driven Supervised Metric Learning (FDSML)}, a pre-training framework that provides models with a structured notion of distance between images.
We first construct \textbf{FineFractalDB}, which substantially increases intra-class diversity through more parameter perturbations, providing the fine-grained relational supervision required for metric learning.
On this dataset, we pre-train models with a \textbf{Parameter-Aware Triplet (PA-Triplet) Loss} 
that assigns positives and negatives according to relative orderings in the image-synthesis parameter space, while discrete class labels only restrict the candidate pool. This objective trains embeddings to respect parameter-derived relative similarity orderings.

Our experimental results demonstrate that FDSML improves transfer beyond classification-based FDSL. FDSML improves over conventional FDSL baselines on several downstream classification benchmarks and, most notably, on image retrieval, highlighting its advantage in transferability. 
It further improves corruption robustness for ResNet-50, with architecture-dependent trade-offs for ViT, and produces qualitative differences in activation and attention patterns.

Our contributions are threefold: (i) we propose \textbf{FDSML}, a synthetic pre-training framework that reformulates FDSL from category recognition to relational comparison; (ii) we develop \textbf{FineFractalDB}, a refined synthetic dataset with substantially higher intra-class diversity for fine-grained relational supervision; and (iii) we introduce the \textbf{PA-Triplet Loss}, which constructs triplets from relative orderings in the generative parameter space, encouraging embeddings that respect parameter-derived similarity orderings.

\section{Related Work}
\label{sec:relatedwork}

\subsection{Learning from Synthetic Data}

Learning visual representations without natural images or manual annotations has emerged as a robust alternative to traditional pre-training, avoiding the high collection costs, restrictive licensing, privacy violations, and implicit biases of large-scale real-image corpora.

\subsubsection{Formula-Driven Supervised Learning (FDSL).}
Formula-driven datasets synthesize both images and supervision from explicit mathematical rules, ensuring full controllability and transparency. FractalDB~\cite{Kataoka_2020_ACCV,Nakashima_Kataoka_Matsumoto_Iwata_Inoue_Satoh_2022} first demonstrated that images generated by Iterated Function Systems (IFS)~\cite{barnsley1988fractals} provide effective pre-training for CNNs and ViTs, and the paradigm has been extended to recursive tilings (TileDB~\cite{Kataoka_2021_ICCV}), contours (RCDB~\cite{9880154}), and sinusoidal structures (VisualAtom~\cite{Takashima_2023_CVPR}), with further explorations into scaling and data efficiency~\cite{Nakamura_2023_ICCV, Nakamura_2024_ECCV}.

However, most FDSL pipelines cast pre-training as classification~\cite{Kataoka_2020_ACCV, Nakashima_Kataoka_Matsumoto_Iwata_Inoue_Satoh_2022, shinoda2023segrcdb, 9880154, Kataoka_2021_ICCV, Takashima_2023_CVPR, Nakamura_2023_ICCV, Nakamura_2024_ECCV}, mapping continuous generation parameters to discrete labels and thereby discarding the neighborhood structure and continuity of the generative process. 
In contrast, we reformulate FDSL as supervised metric learning, exploiting parameter-space relations as ordinal supervision to learn more structured embeddings.

\subsubsection{Data-Free Procedural Programs and Stochastic Generative Processes.}
Representation learning is also possible from stochastic processes and procedural programs without any real-world data, including corpora derived from shader-like programs~\cite{baradad2022procedural} and stochastic occlusion models such as Dead Leaves~\cite{lee2001occlusion} and its derivatives~\cite{baradad2021learning}. 
These methods, however, typically rely on implicit supervision or pretext tasks and do not necessarily provide a continuous, interpretable parameter space for fine-grained geometric control.

\subsubsection{Model-Generated Synthetic Data (Data-Dependent Generators).}
Another line of work builds synthetic datasets with modern generative models such as diffusion models~\cite{rombach2022high} for dataset cloning or augmentation~\cite{trabucco2023effective,wu2023datasetdm}: synthetic ``ImageNet clone'' data yields transferable representations, and diffusion-based augmentation improves supervised training~\cite{sariyildiz2023fake, azizi2023synthetic}, while text-to-image generators enable large corpora for representation learning, such as StableRep~\cite{tian2023stablerep} and learning-from-models paradigms~\cite{tian2024learning}. However, these generators are trained on large real-image corpora and may inherit their biases. In contrast, we focus on formula-driven synthetic data independent of real-image-trained generators and explore how continuous generative factors can shape embedding geometry.

\subsection{Metric Learning and Image Retrieval as Representation Evaluation}
\label{sec:rw_dml_retrieval}

Deep metric learning (DML) learns an embedding space where distances reflect semantic similarity, and image retrieval is a canonical downstream task for diagnosing whether the learned geometry is meaningful. Retrieval systems have progressed from local-feature pipelines~\cite{lowe2004distinctive, o2011introduction, uchida2016image} to deep global/local representations~\cite{ong2017siamesenetworkdeepfishervector, 8237636, cao2020unifying, 10378372, 7301272} and learned re-ranking modules~\cite{fwtan-instance-2021, zhang2020understandingimageretrievalreranking, Zhu_2023_CVPR}. We use retrieval performance as a probe of whether synthetic pre-training yields embeddings with faithful neighborhood structure, beyond classification accuracy alone.

Core DML objectives include contrastive loss~\cite{hadsell2006dimensionality} and triplet-based formulations~\cite{Schroff_2015_CVPR, Song_2016_CVPR, hoffer2015deep}, often paired with sample mining strategies that strongly affect convergence and embedding quality~\cite{Wu_2017_ICCV, hermans2017defense}. Structured objectives such as multi-similarity loss and ranked list loss exploit dense pairwise relations~\cite{wang2019multi, wang2021rankedlistlossdeep}, and proxy-based methods improve scalability by replacing explicit triplet enumeration with learnable class proxies~\cite{Movshovitz-Attias_2017_ICCV, Kim_2020_CVPR}. In parallel, self-supervised contrastive learning (e.g., SimCLR~\cite{chen2020simple,chen2020big}, MoCo~\cite{he2020momentum, chen2020improved, chen2021empirical}, BYOL~\cite{grill2020bootstrap}) shows that instance discrimination yields strong representations without labels, and recent methods explore richer geometries such as hyperbolic and hierarchy-aware metric learning~\cite{kim2023hier, ermolov2022hyperbolic}.

Most prior DML methods construct similarity constraints from discrete class labels (supervised DML)~\cite{NEURIPS2020_d89a66c7} or image augmentations (self-supervised contrastive learning).
In contrast, our approach exploits continuous generative factors in formula-driven synthetic data: we derive relative similarity orderings from distances in the IFS parameter space and train a parameter-aware triplet objective, encouraging the learned embedding geometry to respect these orderings rather than only discrete class partitions.

\section{Method}
\label{sec:method}

We present a metric learning framework tailored to the continuous generative nature of fractal images.
We first revisit the Iterated Function System (IFS) underlying fractal generation, then reformulate pre-training as a metric learning problem with the \textbf{Parameter-Aware Triplet (PA-Triplet) Loss}, and finally describe a refined dataset design that enhances structural diversity and stability.

\subsection{Background: Fractal Generation with Iterated Function System}
\label{sec:method_ifs}

We briefly review IFS, a framework for generating stochastic fractals widely used in synthetic datasets such as FractalDB~\cite{Kataoka_2020_ACCV}.

Formally, an IFS on a complete metric space $(X, d)$ is a finite set of contraction mappings with selection probabilities,
$\mathcal{S} = \left\{ (w_k, \pi_k) \mid k = 1, \dots, N \right\}$,
where each $w_k : X \to X$ is a contraction mapping and $\pi_k > 0$ satisfies $\sum_{k} \pi_k = 1$.
FractalDB employs an affine IFS, in which each transformation $w_k$ is an affine map in $\mathbb{R}^2$:
\begin{equation}
    w_k(\boldsymbol{x}) =
    \begin{bmatrix}
        a_k & b_k \\
        c_k & d_k
    \end{bmatrix}
    \boldsymbol{x}
    +
    \begin{bmatrix}
        e_k \\
        f_k
    \end{bmatrix},
\end{equation}
where $\boldsymbol{x} \in \mathbb{R}^2$. For notational convenience, we denote the affine parameters by $\boldsymbol{\psi}_k = (a_k, b_k, c_k, d_k, e_k, f_k)$.

For generation, the number of transformations $N$ and parameters $\boldsymbol{\psi}_k\sim\mathcal{U}[-1,1]$ are sampled, with $\{\pi_k\}_{k=1}^N$ set proportional to the absolute determinant of each affine matrix.
Starting from $\boldsymbol{x}_0=(0,0)$, points are updated as $\boldsymbol{x}_{i+1}=w_{k_i}(\boldsymbol{x}_i)$, where $k_i$ is drawn from $\{1,\dots,N\}$ with probability $\pi_{k_i}$, producing an orbit that converges to the IFS attractor.

Since the affine parameters $\{\boldsymbol{\psi}_k\}$ determine the attractor geometry, they provide a direct parameterization of fractal shapes, enabling labeled synthetic datasets where each image is associated with its generating parameters. 

\subsection{FDSML: Formula-Driven Supervised Metric Learning}
\label{sec:method_fdsml}

Previous works have explored pre-training from procedurally generated data under the paradigm of Formula-Driven Supervised Learning (FDSL), where labels are derived directly from the parameters of generative formulas.
In the case of synthetic fractal datasets~\cite{Kataoka_2020_ACCV, Anderson_2022_WACV, Nakamura_2023_ICCV, Nakamura_2024_ECCV}, this paradigm is typically realized as a classification problem.
In these approaches, the continuous parameter space is discretized into a finite set of classes, inevitably discarding fine-grained geometric relationships among fractals.
Moreover, classification objectives encourage the model to optimize decision boundaries, which may promote memorization of instance-specific patterns~\cite{zhang2017understanding, ICML-2017-ArpitJBKBKMFCBL}.
However, since individual fractal shapes lack intrinsic semantic meaning, memorizing particular instances does not necessarily result in representations that generalize well to downstream tasks.

To better exploit the continuous nature of fractal generation, we reformulate pre-training as a supervised metric learning problem, which we term \textbf{Formula-Driven Supervised Metric Learning (FDSML)}.
Unlike classification-based FDSL, which enforces discrete category boundaries, 
FDSML models relative orderings among samples.
Because fractals are generated from a continuous parameter space where structural changes can occur gradually, 
the supervision signal naturally lies in which of two candidates is closer to a given reference, rather than in class membership.
By learning to respect such relative orderings derived from the underlying generative formulas, FDSML encourages representations that are less biased toward arbitrary class partitions and more transferable to downstream tasks.

\subsection{Parameter-Aware Triplet Loss}
\label{sec:method_pa-triplet}

\begin{figure*}[!t]
    \begin{center}
        \centering
        \includegraphics[width=1.0\textwidth]{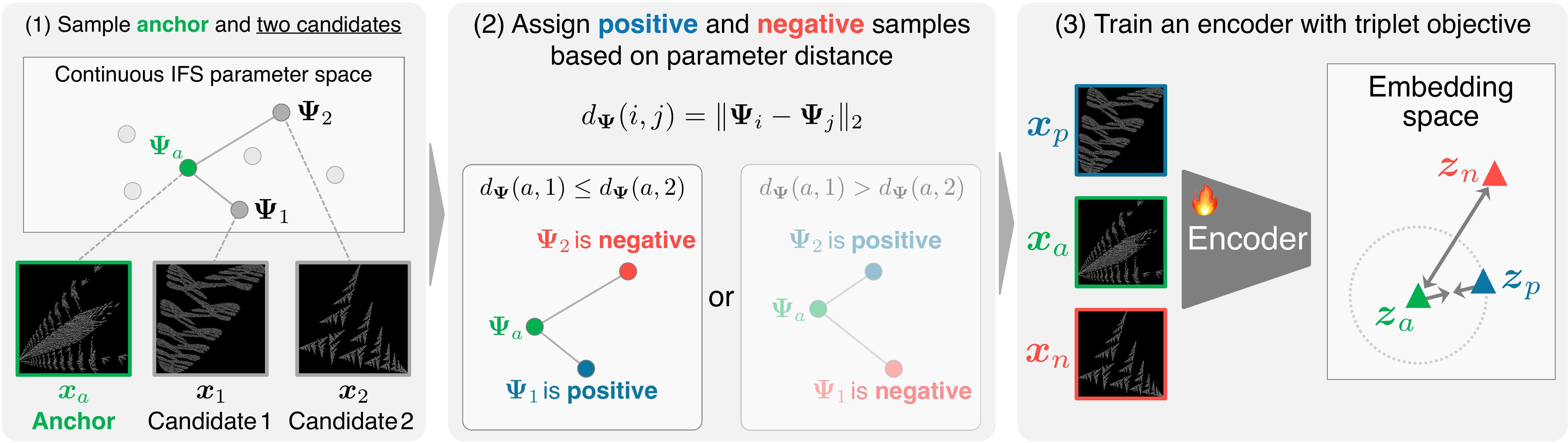}
        \caption{
            Overview of the proposed framework with the Parameter-Aware Triplet (PA-Triplet) Loss.
            (1) An anchor image $\boldsymbol{x}_a$ and two candidate images $\boldsymbol{x}_1$ and $\boldsymbol{x}_2$ are sampled from the continuous IFS parameter space together with their associated parameter descriptors $\boldsymbol{\Psi}_{a}$, $\boldsymbol{\Psi}_{1}$, and $\boldsymbol{\Psi}_{2}$.
            (2) The candidates are assigned as positive or negative according to their \emph{parameter-space} distances to the anchor, where $d_{\boldsymbol{\Psi}}(i,j)=\|\boldsymbol{\Psi}_{i}-\boldsymbol{\Psi}_{j}\|_2$.
            (3) The resulting triplet $(\boldsymbol{x}_a,\boldsymbol{x}_p,\boldsymbol{x}_n)$ is encoded by a feature extractor into embeddings $(\boldsymbol{z}_a,\boldsymbol{z}_p,\boldsymbol{z}_n)$, and the network is optimized so that the \emph{embedding-space} distances reflect the relative ordering induced by the underlying IFS parameter space.
        }
        \label{fig:chap_03_method/pa_triplet}
        \vspace{-1mm}
    \end{center}
\end{figure*}

We adopt a standard triplet-based metric learning objective~\cite{hoffer2015deep, Schroff_2015_CVPR}.
Let $g_{\theta} : \mathcal{X} \subset \mathbb{R}^d \to \mathbb{R}^m$ denote a feature extractor that maps an input image $\boldsymbol{x}$ to an embedding vector.
Given an anchor $\boldsymbol{x}_a$, a positive $\boldsymbol{x}_p$, and a negative $\boldsymbol{x}_n$, the triplet loss is defined as
\begin{equation}
    \resizebox{\linewidth}{!}{$
        \mathcal{L}_{\mathrm{triplet}} =
        \max\Bigl(
        0,\;
        \lVert g_{\theta}(\boldsymbol{x}_a) - g_{\theta}(\boldsymbol{x}_p) \rVert_2
        -
        \lVert g_{\theta}(\boldsymbol{x}_a) - g_{\theta}(\boldsymbol{x}_n) \rVert_2
        + \alpha
        \Bigr)
    $},
    \label{eq:triplet_loss}
\end{equation}
where $\alpha > 0$ is a margin hyperparameter.

In conventional supervised metric learning~\cite{Schroff_2015_CVPR, NEURIPS2020_d89a66c7, Movshovitz-Attias_2017_ICCV}, positives and negatives are determined by discrete class labels.
In contrast, we construct triplets from distances defined in the continuous parameter space of the IFS (\cref{fig:chap_03_method/pa_triplet}).

Given a parameter set $\{(\boldsymbol{\psi}_k, \pi_k)\}_{k=1}^{N}$ that generates an image $\boldsymbol{x}_i$,
we summarize the generative process by the expectation of the affine parameters,
$\boldsymbol{\Psi}_{i} = \sum_{k=1}^{N} \pi_k \, \boldsymbol{\psi}_k \in \mathbb{R}^6$,
a compact descriptor summarizing the weighted affine parameters of the underlying IFS.

We then define a parameter-space distance $d_{\boldsymbol{\Psi}}(i,j) = \| \boldsymbol{\Psi}_{i} - \boldsymbol{\Psi}_{j} \|_{2}$.
For an anchor $\boldsymbol{x}_a$ with descriptor $\boldsymbol{\Psi}_{a}$ and two candidates $\boldsymbol{x}_1$ and $\boldsymbol{x}_2$,
$\boldsymbol{x}_1$ is treated as the positive and $\boldsymbol{x}_2$ as the negative whenever $d_{\boldsymbol{\Psi}}(a,1) \leq d_{\boldsymbol{\Psi}}(a,2)$.
Note that this formulation uses $d_{\boldsymbol{\Psi}}$ only ordinally, to decide which candidate is closer, rather than as a perceptual distance, training the embedding to respect relative orderings instead of categorical separability.
We refer to this formulation as \textbf{Parameter-Aware Triplet (PA-Triplet) Loss}, as triplet construction is explicitly guided by distances in the continuous parameter space.

\begin{figure}[!t]
    \begin{center}
        \centering
        \includegraphics[width=1.0\linewidth]{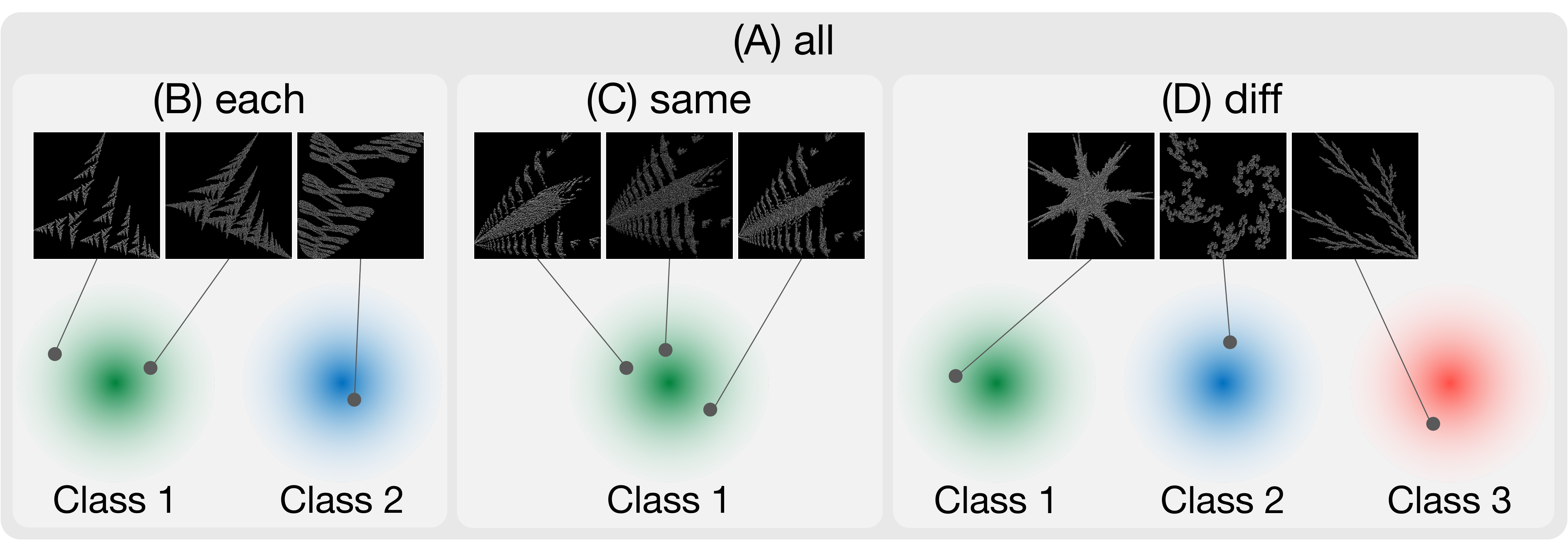}
        \caption{Conceptual illustration of the coarse filtering strategy used to construct candidate triplets.
            \textbf{(A)} \textbf{all}: Two candidates are randomly drawn from the entire dataset.
            \textbf{(B)} \textbf{each}: One candidate is drawn from the same discrete class as the anchor, and the other from a different class.
            \textbf{(C)} \textbf{same}: Both candidates are drawn from the same discrete class as the anchor.
            \textbf{(D)} \textbf{diff}: 
            Both candidates are drawn from classes other than the anchor's class. Discrete labels only restrict the candidate pool; the final positive and negative assignments are determined by parameter-space distance orderings.
            }
        \label{fig:chap_03_method/filtering_triplet_candidates}
        \vspace{-1mm}
    \end{center}
\end{figure}

\begin{figure*}[!t]
    \begin{center}
        \centering
        \includegraphics[width=1.0\textwidth]{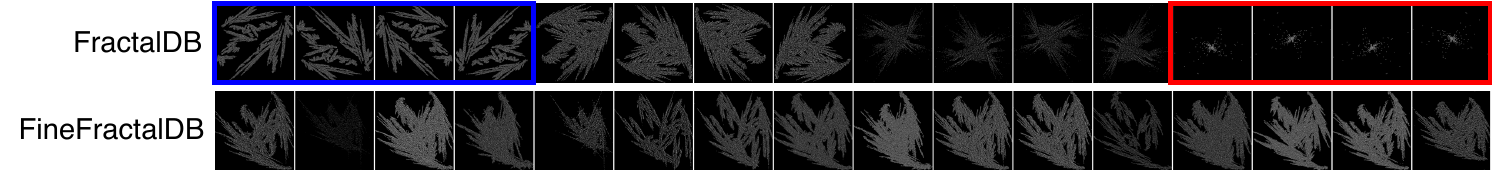}
        \caption{
            Comparison of FractalDB~\cite{Kataoka_2020_ACCV} and FineFractalDB. FractalDB-1k generates 1,000 instances per class from 25 parameter perturbations with flips and rendering variations. The \blue{blue box (left)} shows samples differing only by flipping (limited intra-class diversity), while the \red{red box (right)} shows degenerate samples from non-contractive transformations. FineFractalDB instead uses independent perturbations with strict contraction constraints, producing richer structural variation and far fewer collapsed patterns.
        }
        \label{fig:chap_03_method/finefractaldb_comparison}
        \vspace{-2mm}
    \end{center}
\end{figure*}

\subsection{Filtering Triplets with Discrete Labels}
\label{sec:method_triplet_filtering}

Triplet mining is crucial in deep metric learning, strongly affecting both convergence and embedding quality~\cite{hermans2017defense, Wu_2017_ICCV}.
In preliminary experiments, we observed that excessively large distances between IFS parameter vectors produce renderings with such substantial structural differences that the similarity relation becomes trivial, providing little useful constraint for learning a meaningful metric structure.

To mitigate this issue, we introduce discrete class labels as an auxiliary signal to restrict the candidate pool for triplet construction. 
Although positive/negative assignments are determined by parameter-space distance orderings, discrete labels provide a coarse structural prior that controls the candidate regime.

The filtering strategies differ in how candidates are constrained before triplet construction (\cref{fig:chap_03_method/filtering_triplet_candidates}).
The \textbf{all} setting imposes no restriction, allowing maximal structural variation; \textbf{each} requires one within-class and one cross-class candidate; \textbf{same} restricts both candidates to the anchor's class, focusing on finer structural differences; and \textbf{diff} draws both from other classes, emphasizing larger variations.
Within each filtered set, positives and negatives are determined by relative distances in the continuous parameter space.
These strategies regulate the difficulty and informativeness of the learned similarity constraints.

\subsection{Fine-Grained Fractal Database}
\label{sec:method_finefractaldb}

To better support metric learning, we refine the fractal dataset design inspired by Anderson and Farrell~\cite{Anderson_2022_WACV}.
While the original FractalDB~\cite{Kataoka_2020_ACCV} is effective for classification-based pre-training, it is less suitable for fine-grained relational supervision for two reasons: each class is generated from only a small number of parameter perturbations, resulting in low intra-class diversity, and affine transformations are not strictly constrained to be contraction mappings, occasionally producing visually collapsed patterns with limited structural information.

To address these issues, we increase the number of parameter perturbations to 1,000 per class, matching the number of instances and substantially increasing intra-class diversity.
We also enforce strict contraction constraints on affine transformations to ensure stable fractal structures.
In addition, we apply rejection sampling based on the pixel occupation rate, discarding samples whose occupied pixel ratio is below 0.05\%.

We refer to this refined dataset as the \textbf{Fine-grained Fractal Database (FineFractalDB)}.
As shown in \cref{fig:chap_03_method/finefractaldb_comparison}, FineFractalDB exhibits greater structural variation and far fewer degraded samples than the original FractalDB,  improving both the stability and diversity of generated fractals and making the dataset better suited for learning variation-sensitive representations through metric learning.

\section{Experiments}
\label{sec:experiments}

\subsection{Pre-training Dataset}
\label{sec:datasets}

For our main FDSML pre-training, we use FineFractalDB (Sec.~\ref{sec:method_finefractaldb}), which contains 1,000 categories with 1,000 synthesized images per class (1 million samples in total). 

\subsection{Downstream Benchmarks}
\subsubsection{Image Classification.}
We report Top-1 accuracy after fine-tuning on diverse natural-image datasets: CIFAR-10/100 (C10, C100)~\cite{krizhevsky2009learning}, ImageNet-1k/100 (IN1k, IN100)~\cite{russakovsky2015imagenet}, Places-30 (P30)~\cite{7968387}, Stanford Cars (Cars)~\cite{krause20133d}, and Flowers-102 (Flowers)~\cite{4756141}.
ImageNet-100 and Places-30 follow the evaluation protocols of the original FractalDB study~\cite{Kataoka_2020_ACCV}: the former uses the predefined 100-class subset for efficient benchmarking, and the latter selects 30 scene categories from Places365~\cite{7968387} to evaluate scene-level recognition.

\subsubsection{Image Retrieval.}
We use the Stanford Online Products (SOP)~\cite{Song_2016_CVPR}, Stanford Cars~\cite{krause20133d}, and In-Shop Clothes Retrieval (In-Shop)~\cite{liuLQWTcvpr16DeepFashion} datasets, reporting Recall@$K$ [\%] with $K \in \{1, 10, 100\}$ for SOP, $\{1, 4, 16\}$ for Cars, and $\{1, 10, 20\}$ for In-Shop.
This evaluation verifies whether semantically similar instances are correctly ranked in nearest-neighbor search within the learned representation space.

\subsubsection{Robustness.}
To investigate robustness under distribution shifts and out-of-distribution (OOD) corruptions, we evaluate our models on the CIFAR-10-C and ImageNet-C benchmarks~\cite{hendrycks2019robustness}.

\subsection{Implementation Details}
\label{sec:impl_details}

\subsubsection{Network Architectures.}
We adopt ResNet-50~\cite{he2016deep}, ViT-Tiny, and ViT-Base~\cite{dosovitskiy2020vit} as backbones.
The final fully connected layer of ResNet-50 is replaced with a $512$-dimensional linear embedding layer.
For ViT, following the MoCo v3~\cite{chen2021empirical} design, the classification head is replaced with a 2-layer MLP projection head (Linear $\to$ BatchNorm $\to$ ReLU $\to$ Linear).
The hidden and output dimensions are $192$ and $128$ for ViT-Tiny, and $768$ and $256$ for ViT-Base.

\subsubsection{Optimization and Training Schedule.}
For ResNet-50, we use SGD~\cite{bottou2010large} with momentum $0.9$ and weight decay $1.0 \times 10^{-4}$; the learning rate is linearly scaled with the effective batch size.
Following the FDSL protocol~\cite{Kataoka_2020_ACCV}, the learning rate is kept constant for the first 40\% of training and then decayed with a cosine schedule, and the model is pre-trained for 134 epochs.
For ViT, we use AdamW~\cite{loshchilov2017decoupled} with a weight decay of $0.05$ and a base learning rate of $1.0 \times 10^{-4}$ for ViT-Tiny and $1.0 \times 10^{-5}$ for ViT-Base, following the linear scaling rule ($\text{base\_lr} \times \text{BatchSize} / 256$)~\cite{chen2021empirical}.
We apply a 10,000-step linear warm-up followed by cosine decay, and pre-train for 536 epochs to account for the slower convergence of transformers on synthetic data.

\begin{table}[t]
    \centering
    \caption{
        Agreement [\%] between the $d_{\boldsymbol{\Psi}}$ and Chamfer distance orderings under each filtering strategy.
        \emph{Label only} assigns the positive by class label ($\dagger$: at chance by definition).
        The value for \textbf{all} is omitted (--), as it would reflect the class composition of the dataset rather than information extractable from the labels.
    }
    \label{tab:agreement}
    \resizebox{0.5\columnwidth}{!}{
        \begin{tabular}{lcc}
            \toprule
            \multirow{2.5}{*}{Filtering} & \multicolumn{2}{c}{Agreement [\%]} \\
            \cmidrule(lr){2-3}
            & $d_{\boldsymbol{\Psi}}$ & Label only \\
            \midrule
            all  & 47.2 & - \\
            each & 89.2 & 89.4 \\
            same & 80.6 & 50.0$^{\dagger}$ \\
            diff & 46.8 & 50.0$^{\dagger}$ \\
            \bottomrule
        \end{tabular}
    }
\end{table}

\section{Experimental Results}
\subsection{Validating the Parameter Descriptor}
\label{sec:agreement}

The PA-Triplet loss uses $d_{\boldsymbol{\Psi}}$ only to rank two candidates relative to an anchor (\cref{sec:method_pa-triplet}).
We therefore validate whether the candidate ranked closer in $d_{\boldsymbol{\Psi}}$ is also the structurally closer one once rendered, using the symmetric Chamfer distance~\cite{chamfer} as the reference measure.
Binarizing two renderings into foreground pixel sets $A$ and $B$, we compute
\begin{equation}
  \resizebox{0.85\linewidth}{!}{$
      {\displaystyle D(A,B) = \frac{1}{2}\left(
        \frac{1}{|A|}\sum_{\mathbf{p}\in A}\min_{\mathbf{q}\in B}\|\mathbf{p}-\mathbf{q}\|_1
        + \frac{1}{|B|}\sum_{\mathbf{q}\in B}\min_{\mathbf{p}\in A}\|\mathbf{q}-\mathbf{p}\|_1
      \right),}
  $}
  \label{eq:chamfer}
\end{equation}
which averages the distance from each foreground pixel to its nearest counterpart in the other set.
Since each attractor is min--max normalized to the image frame before rasterization, $D$ reflects differences in shape rather than global position or scale.
We verify in the supplementary material that $D$ responds consistently to controlled generative perturbations and that $d_{\boldsymbol{\Psi}}$ tracks it closely, establishing $D$ as a reliable reference for structural difference.

\begin{table}[!t]
\centering
\caption{Top-1 classification accuracy (\%) on ResNet-50.}
\label{tab:classification_resnet}
\scriptsize
\setlength{\tabcolsep}{3pt}
\resizebox{\columnwidth}{!}{%
\begin{tabular}{llccccc}
\toprule
PT Dataset & PT Type & C10 & C100 & IN100 & IN1k & P30 \\
\midrule
Scratch & - & 87.60 & 62.70 & 72.42 & 75.47 & 73.50 \\
ImageNet-1k & SL & 96.96 & 84.48 & - & - & 80.33 \\
\midrule
FractalDB-1k & FDSL & 90.50 & \textbf{70.21} & 79.42 & 70.43 & 76.33 \\
\rowcolor{ldblue}
FineFractalDB & FDSML (all) & \underline{90.79} & 68.49 & 79.62 & 71.37 & 76.07 \\
\rowcolor{ldblue}
FineFractalDB & FDSML (each) & 90.53 & 67.46 & 78.80 & 71.21 & \textbf{77.53} \\
\rowcolor{ldblue}
FineFractalDB & FDSML (same) & \textbf{91.47} & \underline{68.74} & \textbf{80.62} & \textbf{72.65} & \underline{77.20} \\
\rowcolor{ldblue}
FineFractalDB & FDSML (diff) & 90.29 & 68.37 & \underline{79.66} & \underline{71.56} & 77.17 \\
\bottomrule
\end{tabular}
}
\end{table}
\begin{table}[t]
\centering
\caption{Top-1 classification accuracy (\%) on ViT-Tiny.}
\label{tab:classification_vittiny}
\scriptsize
\setlength{\tabcolsep}{3pt}
\resizebox{\columnwidth}{!}{%
\begin{tabular}{llccccccc}
\toprule
PT Dataset & PT Type & C10 & C100 & IN100 & IN1k & P30 & Cars & Flowers \\
\midrule
Scratch & - & 76.47 & 58.02 & 70.08 & 71.46 & 81.17 & 21.30 & 48.74 \\
ImageNet-1k & SL & 97.92 & 85.65 & - & - & 81.48 & 91.26 & 93.70 \\
\midrule
FractalDB-1k & FDSL & 96.65 & 81.18 & 84.40 & 72.40 & 80.46 & 77.01 & \underline{73.99} \\
\rowcolor{ldblue}
FineFractalDB & FDSML (all) & 97.71 & \underline{81.73} & \underline{88.47} & 73.68 & 80.25 & 80.83 & 66.99 \\
\rowcolor{ldblue}
FineFractalDB & FDSML (each) & \textbf{98.00} & \textbf{82.08} & \textbf{88.90} & 73.71 & 80.35 & \textbf{83.17} & \textbf{74.21} \\
\rowcolor{ldblue}
FineFractalDB & FDSML (same) & 97.71 & 81.68 & 88.28 & \underline{73.80} & \textbf{80.58} & \underline{81.72} & 69.23 \\
\rowcolor{ldblue}
FineFractalDB & FDSML (diff) & \underline{97.72} & 80.68 & 88.25 & \textbf{73.97} & \underline{80.49} & 79.84 & 66.96 \\
\bottomrule
\end{tabular}%
}
\end{table}
\begin{table}[t]
    \centering
    \caption{Top-1 classification accuracy (\%) on ViT-Base.}
    \label{tab:classification_vitbase}

    \resizebox{0.8\linewidth}{!}{
        \begin{tabular}{llccc}
            \toprule
            PT Dataset & PT Type & C10 & C100 & IN100 \\
            \midrule
            Scratch & -- & 90.03 & 66.83 & 87.64 \\
            ImageNet-1k & SL & 98.49 & 85.73 & -- \\
            \midrule
            FractalDB-1k & FDSL & \underline{98.19} & \textbf{83.68} & \underline{89.59} \\
            \rowcolor{ldblue}
            FineFractalDB & FDSML (all)  & 97.96 & 81.79 & 89.28 \\
            \rowcolor{ldblue}
            FineFractalDB & FDSML (each) & \textbf{98.23} & \underline{83.30} & \textbf{90.09} \\
            \rowcolor{ldblue}
            FineFractalDB & FDSML (same) & 98.12 & 82.19 & 89.52 \\
            \rowcolor{ldblue}
            FineFractalDB & FDSML (diff) & 98.15 & 82.77 & 89.41 \\
            \bottomrule
        \end{tabular}
    }
\end{table}

For triplets sampled under each filtering strategy, \Cref{tab:agreement} reports how often the candidate closer in $d_{\boldsymbol{\Psi}}$ is also closer in $D$, which is exactly the information PA-Triplet uses to assign positives and negatives.
\textbf{each} and \textbf{same} reach 89.2\% and 80.6\%, respectively, against a chance level of 50\%, confirming a valid similarity signal in these regimes.
To isolate the descriptor's contribution, we compare with assigning the positive by class label alone.
Under \textbf{each}, this label rule attains 89.4\%, and the descriptor picks the same positive in 99.6\% of triplets, so its supervision is effectively equivalent to that of the labels.
Under \textbf{same}, no label rule exists because all three images share a class, yet the descriptor attains 80.6\%, a signal that class labels cannot provide even in principle.
In contrast, \textbf{all} and \textbf{diff} fall below chance, at 47.2\% and 46.8\%, respectively.
Because $\boldsymbol{\Psi}$ compresses each IFS into a six-dimensional expectation, dissimilar transformation sets can map to nearby descriptors, mainly across class boundaries.
These results indicate that the reliability of the descriptor ordering depends strongly on the candidate regime: \textbf{same} provides meaningful within-class ordinal structure, whereas comparisons among unrelated classes under \textbf{diff} are unreliable.

\begin{table}[t]
\centering
\caption{Ablation study on pre-training configurations. Top-1 accuracy (\%) of ViT-Tiny on downstream datasets.}
\label{tab:ablation}
\resizebox{0.8\linewidth}{!}{
  \begin{tabular}{llccc}
    \toprule
    PT Type & PT Dataset & C10 & C100 & IN100 \\
    \midrule
    FDSL         & FractalDB-1k  & 96.65 & 81.18 & 84.40 \\
    FDSL         & FineFractalDB & 96.11 & 77.52 & 84.32 \\
    \midrule
    FDSML (each) & FractalDB-1k  & \underline{97.77} & \underline{81.72} & 87.93 \\
    \rowcolor{ldblue}
    FDSML (each) & FineFractalDB & \textbf{98.00} & \textbf{82.08} & \textbf{88.90} \\
    \midrule
    FDSML (same) & FractalDB-1k  & 97.16 & 79.76 & 87.23 \\
    \rowcolor{ldblue}
    FDSML (same) & FineFractalDB & 97.71 & 81.68 & \underline{88.28} \\
    \bottomrule
  \end{tabular}
}
\end{table}
\begin{table*}[!t]
\centering
\caption{Classification Error (CE $\downarrow$, \%) on ImageNet-C. Each entry reports the average CE over corruption types within each group.}
\label{tab:corruption}

\begin{subtable}{0.495\linewidth}
\centering
\caption{ResNet-50}

\resizebox{\linewidth}{!}{
\begin{tabular}{llccccc}
\toprule
PT Dataset & PT Type & Noise & Blur & Weather & Digital & Geometric \\
\midrule
FractalDB-1k & FDSL & 86.16 & 74.82 & 71.94 & 60.36 & 61.15 \\
\rowcolor{ldblue}
FineFractalDB & FDSML (all)  & \underline{84.06} & \textbf{72.55} & \textbf{68.50} & 59.40 & \underline{59.85} \\
\rowcolor{ldblue}
FineFractalDB & FDSML (each) & 85.22 & 73.25 & 70.06 & \underline{58.97} & 61.23 \\
\rowcolor{ldblue}
FineFractalDB & FDSML (same) & \textbf{83.62} & \underline{72.87} & \underline{68.81} & \textbf{57.46} & \textbf{59.13} \\
\rowcolor{ldblue}
FineFractalDB & FDSML (diff) & 84.56 & 73.41 & 69.55 & 59.27 & 61.10 \\
\bottomrule
\end{tabular}
}

\end{subtable}
\hfill
\begin{subtable}{0.495\linewidth}
\centering
\caption{ViT-Tiny}
\resizebox{\linewidth}{!}{
\begin{tabular}{llccccc}
\toprule
PT Dataset & PT Type & Noise & Blur & Weather & Digital & Geometric \\
\midrule
FractalDB-1k & FDSL         & \textbf{55.31} & 60.78 & 50.84 & \textbf{41.94} & 48.49 \\
\rowcolor{ldblue}
FineFractalDB & FDSML (all)  & 60.26 & 60.78 & \underline{48.15} & 43.96 & \textbf{47.90} \\
\rowcolor{ldblue}
FineFractalDB & FDSML (each) & \underline{58.69} & \underline{60.13} & 48.34 & \underline{43.41} & 48.40 \\
\rowcolor{ldblue}
FineFractalDB & FDSML (same) & 59.26 & \textbf{60.06}  & \textbf{47.64} & 43.55 & \underline{47.97} \\
\rowcolor{ldblue}
FineFractalDB & FDSML (diff) & 59.47 & 60.54 & 48.22 & 43.88 & 48.29  \\
\bottomrule
\end{tabular}
}

\end{subtable}

\end{table*}

\subsection{Downstream Classification}
We evaluate the transferability of the learned representations on several downstream benchmarks.
Tables~\ref{tab:classification_resnet}, \ref{tab:classification_vittiny}, and~\ref{tab:classification_vitbase} report Top-1 accuracy for ResNet-50, ViT-Tiny, and ViT-Base, respectively.

For ResNet-50 (\Cref{tab:classification_resnet}), while natural-image pre-training (ImageNet-1k) remains the most potent baseline,
our distance-based synthetic pre-training outperforms conventional cross-entropy pre-training on FractalDB-1k on most benchmarks, while remaining competitive on the rest.
In particular, FineFractalDB with FDSML (same) achieves 91.47\% on CIFAR-10 and 72.65\% on ImageNet-1k, 
indicating that distance-based supervision provides more transferable features than classification-based synthetic pre-training.
The strong performance of FDSML (same) is consistent with \Cref{sec:agreement}, 
where \textbf{same} provides parameter-derived orderings that discrete class labels cannot express.

For ViT-Tiny (\Cref{tab:classification_vittiny}), similar trends hold: our metric-based approach improves ImageNet-100 accuracy to 88.90\%, a substantial gain over the 84.40\% of cross-entropy pre-training on FractalDB-1k, and reaches 98.00\% on CIFAR-10, slightly exceeding ImageNet-1k supervised pre-training (97.92\%). These results indicate that the metric-learning reformulation transfers beyond convolutional architectures and, as shown in \Cref{sec:ablation}, accounts for a substantial share of these gains.

\Cref{tab:classification_vitbase} shows that these trends extend to ViT-Base.
Although the gap between FDSL and FDSML narrows relative to ViT-Tiny, FDSML (each) achieves the strongest overall performance, with the highest accuracy on CIFAR-10 and ImageNet-100 while remaining competitive on CIFAR-100.
The PA-Triplet objective thus remains effective even for larger models.

\subsubsection{Ablation Studies.}
\label{sec:ablation}

\Cref{tab:ablation} decomposes the contributions of the pre-training objective and
dataset design with ViT-Tiny.
With the dataset fixed to FractalDB-1k, replacing cross-entropy with
PA-Triplet improves all three benchmarks, isolating the effect of the
metric-learning reformulation from any change in the data.
Under FDSML, FineFractalDB yields gains for both \textbf{each}
and \textbf{same}. In contrast, under FDSL, FineFractalDB provides no
gain over FractalDB-1k and degrades performance on
CIFAR-100.
FineFractalDB increases intra-class variation, but cross-entropy assigns
all instances within a class the same label and therefore does not
directly exploit this variation. PA-Triplet instead uses relative
orderings among samples to construct relational supervision from such
variation. These results support the complementarity of the refined
dataset and the relational objective.

\subsection{Robustness Analysis}
\label{sec:Robustness Evaluation}
We evaluate robustness against common visual corruptions on ImageNet-C.
Table~\ref{tab:corruption} reports the Classification Error averaged within five semantic groups.

For ResNet-50, our distance-based pre-training outperforms the classification-based synthetic baseline (FDSL) across all five corruption categories, with FDSML (same) achieving the best performance in three of them.

For ViT-Tiny, the results reveal a more nuanced trade-off: our method consistently improves resilience against structural and environmental distortions, achieving a 6.3\% relative improvement in the Weather category (47.64 vs.\ 50.84), while degrading in the Noise (58.69 for FDSML (each) vs.\ 55.31) and Digital (43.41 vs.\ 41.94) categories.
These findings suggest that PA-Triplet can improve robustness to structural and environmental distortions, while its relationship to high-frequency noise remains architecture-dependent and requires dedicated frequency analysis.
Detailed CIFAR-10-C evaluations are provided in the supplementary material.

\subsection{Image Retrieval}
\label{sec:Image Retrieval Evaluation}

\begin{table}[!tb]
    \caption{Recall@K [\%] on Stanford Online Products (SOP)~\cite{Song_2016_CVPR}, Stanford Cars~\cite{krause20133d}, and In-Shop Clothes~\cite{liuLQWTcvpr16DeepFashion} using a ResNet-50 backbone under different pre-training strategies.}
    \label{tab:retrieval-recall}
    \centering
    \resizebox{\linewidth}{!}{
        \setlength{\tabcolsep}{3.2pt}
        \normalsize
        \begin{tabular}{llccccccccc}
            \toprule
            \multirow{2}{*}{PT Dataset}
            & \multirow{2}{*}{PT Type}
            & \multicolumn{3}{c}{SOP}
            & \multicolumn{3}{c}{Stanford Cars}
            & \multicolumn{3}{c}{In-Shop} \\
            \cmidrule(lr){3-5} \cmidrule(lr){6-8} \cmidrule(lr){9-11}
            &
            &
            1 & 10 & 100
            & 1 & 4 & 16
            & 1 & 10 & 20 \\
            \midrule
            Scratch        & -                & 70.4 & 83.1 & 90.9                            & 40.9 & 61.5 & 80.4                                     & 87.2 & 95.7 & 96.9 \\
            ImageNet-1k    & SL               & 79.1 & 91.0 & 96.3                            & 87.7 & 95.6 & 98.6                                     & 91.1 & 97.8 & 98.4 \\
            \midrule
            FractalDB-1k   & FDSL             & 65.6 & 81.3 & 91.0                            & 56.2 & 76.3 & 89.8                                     & 87.3 & 95.5 & 96.6 \\
            \rowcolor{ldblue}
            FineFractalDB  & FDSML (all)      & \underline{73.9} & 86.3 & 93.3                            & \underline{77.8} & \underline{89.5} & \underline{95.4}                                     & \textbf{89.2} & \underline{96.9} & \textbf{97.9} \\
            \rowcolor{ldblue}
            FineFractalDB  & FDSML (each)     & \textbf{74.0} & \underline{86.5} & \underline{93.4}                   & 76.2 & 88.6 & 95.2                                     & \textbf{89.2} & 96.7 & \underline{97.8} \\
            \rowcolor{ldblue}
            FineFractalDB  & FDSML (same)     & \textbf{74.0} & \textbf{86.6} & \textbf{93.5} & \textbf{78.4} & \textbf{90.2} & \textbf{96.0}          & \textbf{89.2} & \textbf{97.0} & \textbf{97.9} \\
            \rowcolor{ldblue}
            FineFractalDB  & FDSML (diff)     & 73.6 & 86.3 & 93.2                            & 75.6 & 88.2 & 95.1                                     & \underline{88.9} & \underline{96.9} & \underline{97.8} \\
            \bottomrule
        \end{tabular}
    }
\end{table}

\begin{figure}[!t]
    \begin{center}
        \centering
        \includegraphics[width=1.0\linewidth]{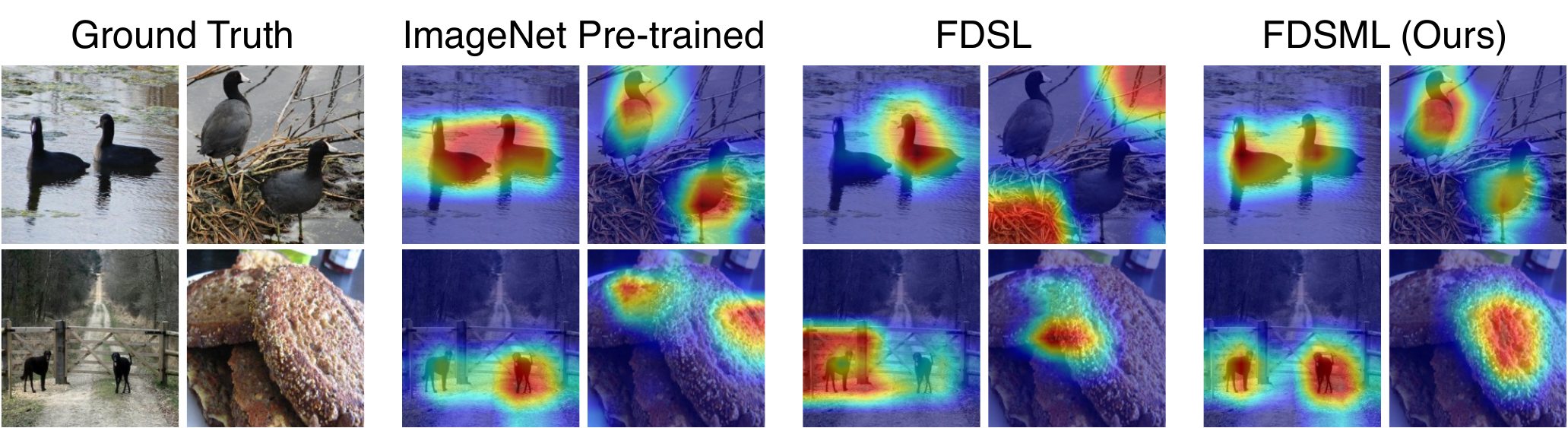}
        \caption{Grad-CAM visualizations for ResNet-50 pre-trained under different methods.}
        \label{fig:chap_03_method/grad_cam}
        \vspace{-1mm}
    \end{center}
\end{figure}

\begin{figure*}[!t]
    \begin{center}
        \centering
        \includegraphics[width=0.9\textwidth]{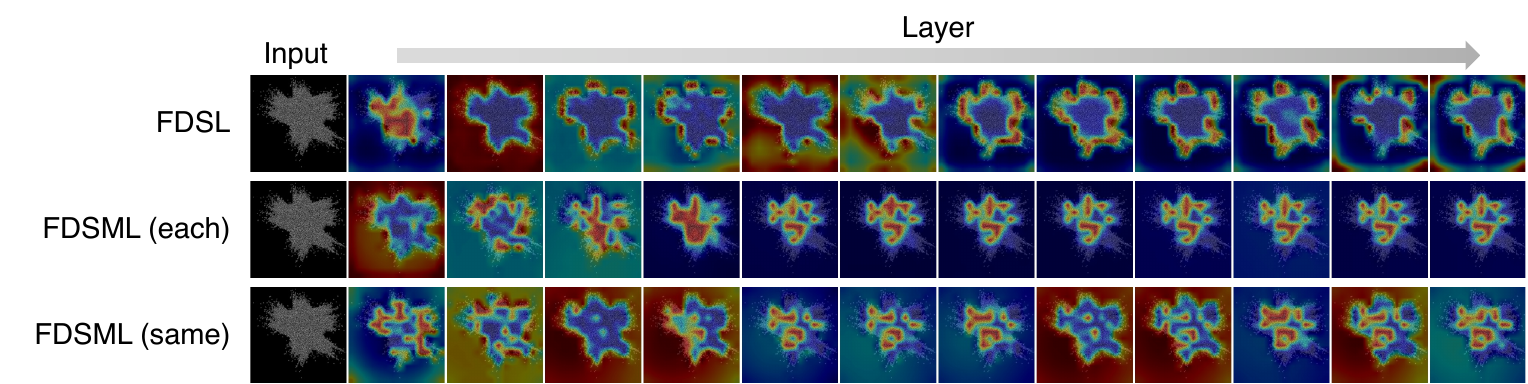}
        \caption{Layer-wise CLS-to-patch attention maps of ViT-Tiny pre-trained with different objectives. Rows correspond to different pre-training strategies and columns correspond to Transformer layers ordered from shallow (left) to deep (right).}
        \label{fig:chap_03_method/layerwise_attn}
        \vspace{-1mm}
    \end{center}
\end{figure*}

\Cref{tab:retrieval-recall} reports Recall@K on SOP, Cars, and In-Shop after fine-tuning with the Proxy-Anchor objective~\cite{Kim_2020_CVPR}.

Across all three datasets, models pre-trained on FineFractalDB with FDSML consistently outperform those trained from scratch and those pre-trained on FractalDB-1k with FDSL, while ImageNet-1k pre-training serves as an upper baseline.
These results suggest that FDSML pre-training better preserves relative similarity structure in the embedding space, yielding representations that transfer more effectively to embedding-based tasks.

Notably, on SOP the FDSL model pre-trained on FractalDB underperforms training from scratch in Recall@1 (65.6 vs.\ 70.4) and Recall@10 (81.3 vs.\ 83.1), indicating that classification-based formula-driven supervision does not necessarily provide a beneficial inductive bias for tasks that rely on embedding geometry.
One possible explanation is that classification objectives enforce categorical separability without modeling relative similarity orderings among samples: while effective for classification, they do not explicitly exploit the ordinal relationships available from the continuous IFS parameter space.

Among the triplet candidate filtering strategies (\cref{sec:method_triplet_filtering}), FDSML (same), which samples both candidates from the anchor's discrete class, performs best on SOP and Stanford Cars and remains on par with the best variant on In-Shop.
These results support the hypothesis that learning to respect the relative ordering of generative variation, rather than merely enforcing categorical separability, leads to representations better suited to embedding-based tasks.

\subsection{Qualitative Analysis}
\label{sec:Qualitative Evaluation}

\subsubsection{Grad-CAM Visualization on ResNet-50.}
\Cref{fig:chap_03_method/grad_cam} presents Grad-CAM~\cite{selvaraju2017gradcam} visualizations for ResNet-50 models pre-trained under different strategies.
Each example shows the input image with its ground-truth label, followed by the Grad-CAM results of (i) the ImageNet-1k pre-trained model, (ii) the FDSL model pre-trained on FractalDB-1k with cross-entropy, and (iii) our FDSML (same) model pre-trained on FineFractalDB with the PA-Triplet objective.

Compared with the FDSL model, our FDSML produces more spatially concentrated activation maps on object-relevant regions, qualitatively closer to those of the ImageNet pre-trained model, whereas FDSL often yields more diffuse and less localized responses.
These observations suggest that PA-Triplet pre-training promotes spatially coherent feature representations, which may contribute to improved transfer performance.

\subsubsection{Visualizing Attention Maps.}
\Cref{fig:chap_03_method/layerwise_attn} presents layer-wise CLS-to-patch attention maps of ViT models pre-trained with different objectives, computed on FractalDB~\cite{Kataoka_2020_ACCV} images across all 12 Transformer layers.
Columns correspond to layers ordered from shallow (left) to deep (right), and rows to pre-training strategies.

The FDSL model concentrates attention primarily along prominent boundaries, whereas the FDSML model also attends to textured and interior regions.
This broader spatial distribution suggests that FDSML captures fine-grained appearance variations in addition to coarse structural cues.
Such representations may be advantageous for downstream tasks that depend on subtle visual distinctions, including fine-grained recognition.
Additional qualitative results are provided in the supplementary material.

\subsection{Limitations}
Our validation is currently limited to IFS-based fractal pre-training.
Although the parameter-aware triplet construction could in principle be instantiated for other FDSL generators such as VisualAtom or TileDB, each generator would require its own similarity definition over the generation parameters, together with dedicated empirical validation.
We therefore leave this extension to future work.
In addition, our results are confined to the one-million-image scale of FineFractalDB, and whether the gains from parameter-aware supervision persist at larger scales remains an open question.
Finally, compared with cross-entropy-based FDSL, FDSML incurs additional overhead for descriptor computation and triplet construction, which we have not yet analyzed in detail.

\section{Conclusion}
\label{sec:conclusion}

We introduced Formula-Driven Supervised Metric Learning (FDSML), a synthetic pre-training framework that leverages continuous generative parameters for representation learning. 
Unlike conventional FDSL, which mainly encourages category recognition, FDSML reformulates synthetic pre-training as a relational learning problem. 
Through the PA-Triplet Loss and FineFractalDB, models are supervised with relative orderings derived from the generative parameter space.
Experiments show that FDSML substantially improves image retrieval over classification-based FDSL while remaining competitive on downstream classification. 
It also improves corruption robustness for ResNet-50, with architecture-dependent trade-offs for ViT, and produces qualitative differences in activation and attention patterns.
These results show that parameter-derived relational supervision can improve transfer beyond category recognition in IFS-based formula-driven synthetic pre-training.

\section{Acknowledgement}

This work was supported by JSPS KAKENHI Grant Numbers JP24H00742 and JP24H00748, JST SPRING Grant Number JPMJSP2128, and computational resources of ABCI~3.0 provided by AIST and AIST Solutions.

{
    \small
    \bibliographystyle{ieeenat_fullname}
    \bibliography{main}
}

\end{document}

\AddToShipoutPicture*{%
  \AtPageUpperLeft{%
    \raisebox{-0.45in}{%
      \makebox[\paperwidth]{%
        \parbox[t]{\textwidth}{\scriptsize\centering\color{gray}
          \copyright~2027 IEEE. Personal use of this material is permitted.
          Permission from IEEE must be obtained for all other uses, in any current or future media, 
          including reprinting/republishing this material for advertising or promotional purposes, creating new collective works, for resale or redistribution to servers or lists, or reuse of any copyrighted component of this work in other works.}}}}}
\maketitle

\begingroup
\renewcommand{\thefootnote}{}%
\footnotetext{$^{\ast}$Equal contribution.}%
\endgroup

\section{Empirical Validation of the Parameter Descriptor}
\label{sec:supp_validation}

The PA-Triplet loss uses the descriptor distance $d_{\boldsymbol{\Psi}}$ only to rank two candidates relative to an anchor.
Section~5.1 of the main paper validates this ordinal supervision directly by measuring how often the candidate ranked closer in $d_{\boldsymbol{\Psi}}$ is also the closer one under the symmetric Chamfer distance $D$~\cite{chamfer} (Eq.~(3) and Table~1 of the main paper).
That analysis presupposes that $D$ itself is a reliable reference for structural difference.
This section supplies the supporting evidence.
\Cref{sec:supp_perturbation} verifies that $D$ responds consistently to controlled perturbations of the generative parameters and that $d_{\boldsymbol{\Psi}}$ tracks this response.
\Cref{sec:supp_triplet_examples} complements the quantitative agreement with qualitative examples of triplets constructed under each filtering strategy.

We adopt $D$ as the reference for two reasons.
First, the renderings are effectively binary, so the binarization threshold is immaterial and $D$ involves no tunable hyperparameters such as a blur width.
Second, unlike learned perceptual metrics (e.g., LPIPS~\cite{zhang2018perceptual}), $D$ assumes no prior over natural images and thus avoids the secondary question of whether such a prior transfers to synthetic fractals.

\subsection{Response to Controlled Generative Perturbations}
\label{sec:supp_perturbation}

Starting from a base IFS code, we rescale one of the six affine coefficients across all maps by a factor $s \in [0.6, 1.4]$ sampled at 33 evenly spaced values including $s{=}1$, and recompute the selection probabilities from the determinant magnitudes.
This is the same mechanism used to synthesize intra-class instances of FineFractalDB, in which one randomly chosen coefficient is rescaled by a factor drawn from the same range.
The sweep therefore probes the type of variation realized in our dataset.
Each perturbed code is rendered and compared with the base rendering by $D$.
By construction, $D$ averages the distance from each foreground pixel to its nearest counterpart in the other rendering.
It is therefore expressed in pixels and indicates how far, on average, the two rendered shapes are displaced from each other.
Because rendering involves stochastic point sampling and random patch patterns, re-rendering identical parameters with different seeds yields a nonzero $D$.
We estimate this rendering noise floor for each sweep from such seed-only re-renderings and use it to separate structural change from rendering variability.

\begin{figure}[t]
  \centering
  \includegraphics[width=\linewidth]{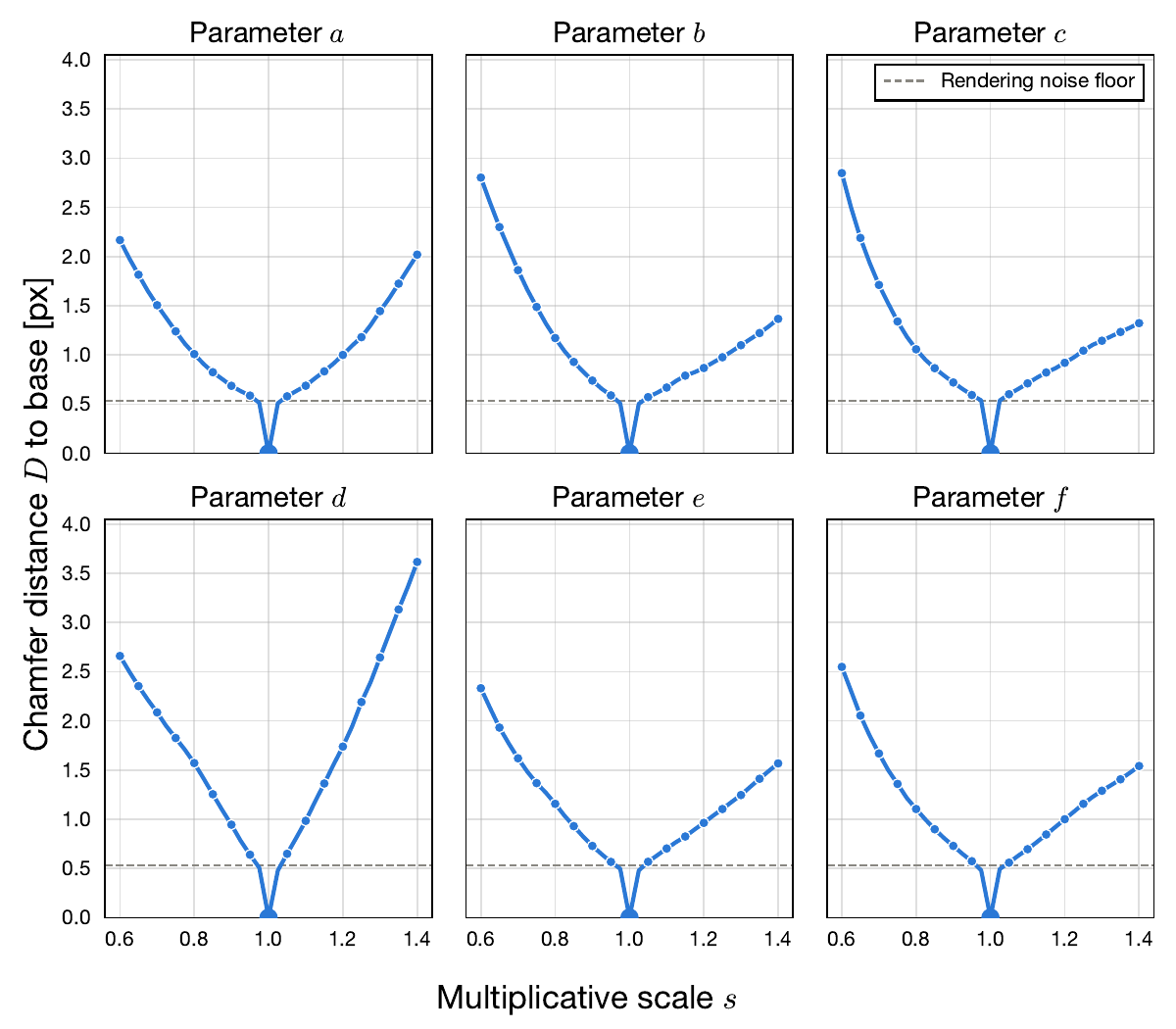}
  \caption{
    Response of the Chamfer distance $D$ to controlled perturbations of a single base IFS parameter set.
    One affine coefficient is rescaled across all maps by a factor $s \in [0.6, 1.4]$ (33 steps), and the selection probabilities are recomputed from the determinant magnitudes.
    Each panel corresponds to one of the six coefficients.
    The dashed line indicates the rendering noise floor, estimated by re-rendering the unperturbed parameters with different random seeds.
    $D$ increases monotonically on both sides of the base in all six directions.
  }
  \label{fig:supp_sweep}
\end{figure}

\begin{figure}[t]
  \centering
  \includegraphics[width=\linewidth]{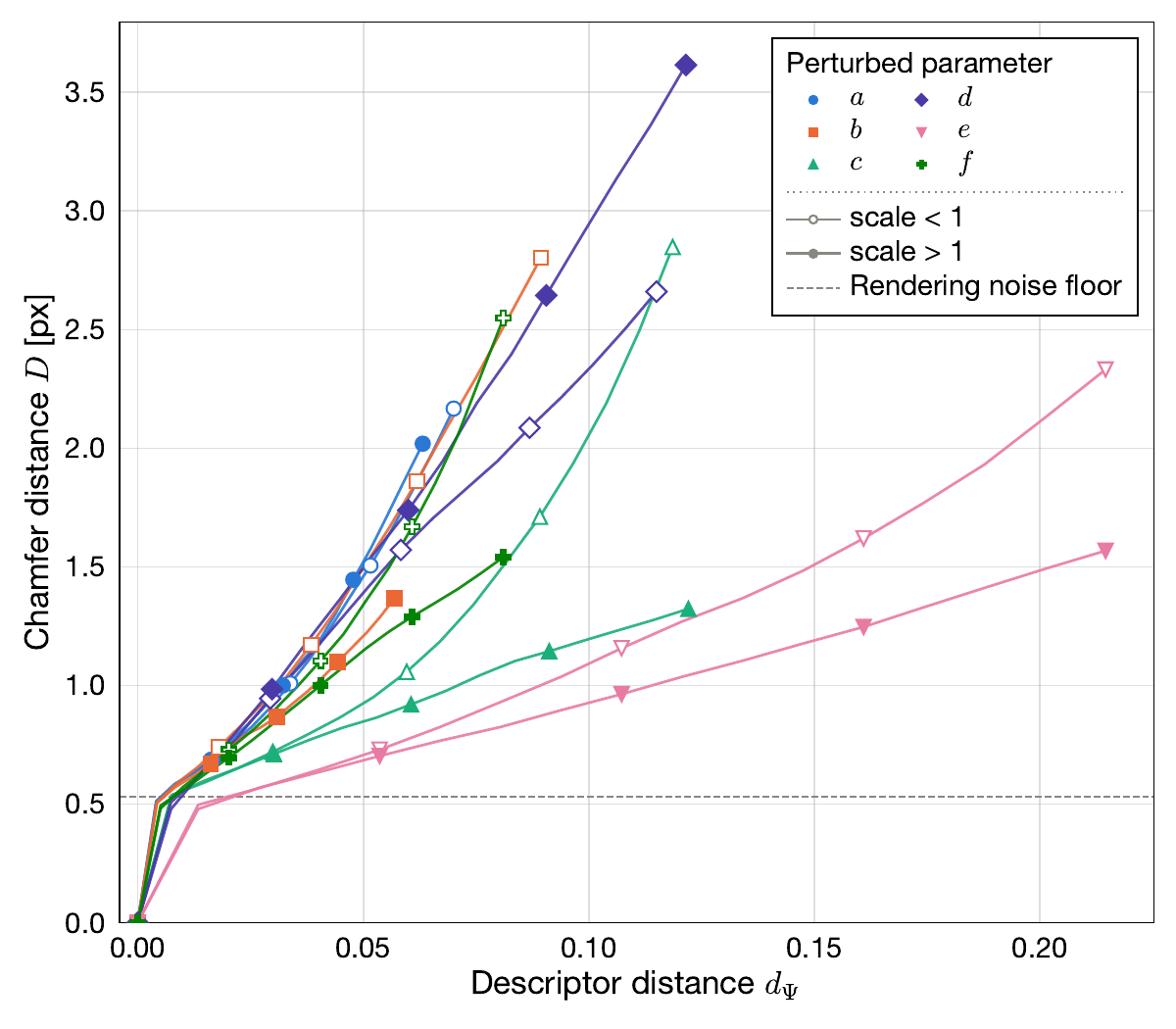}
  \caption{
    Chamfer distance $D$ versus descriptor distance $d_{\boldsymbol{\Psi}}$ for the perturbation sweeps of \cref{fig:supp_sweep}, colored by the perturbed coefficient.
   Each sweep contributes two branches corresponding to shrinking ($s{<}1$) and enlarging ($s{>}1$) perturbations.
   The rank correlation is nearly perfect within each direction, whereas the slope differs across directions, so $d_{\boldsymbol{\Psi}}$ orders structural change reliably but is not calibrated in magnitude across directions.
  }
  \label{fig:supp_scatter}
\end{figure}

\Cref{fig:supp_sweep} illustrates the response of $D$ for a single base IFS parameter set.
In all six directions, $D$ increases monotonically on both sides of $s{=}1$, and only the immediate neighborhood of the base falls within the noise floor of 0.53\,px.
The largest displacement per direction reaches 2.2--3.6\,px, four to seven times the noise floor.
\Cref{fig:supp_scatter} plots $D$ against $d_{\boldsymbol{\Psi}}$ for the same sweeps.
Within each direction, the Spearman rank correlation coefficient $\rho$ between $d_{\boldsymbol{\Psi}}$ and $D$ is nearly perfect ($\rho \geq 0.94$), whereas pooling all directions yields $\rho = 0.831$ because the slope of the response differs across directions by roughly a factor of four for this IFS.
The descriptor therefore orders structural change reliably along each direction but is not calibrated in magnitude across directions, which matches the strictly ordinal use of $d_{\boldsymbol{\Psi}}$ in PA-Triplet (Sec.~3.3 of the main paper).

To move beyond a single example, we repeat the sweep for 50 base IFS codes taken from distinct FineFractalDB classes and for each of the six coefficients, yielding 300 sweeps in total.
Among them, 198 sweeps remain contractive over the entire range, in the sense that the largest singular value of every affine map stays below one.
In 91.4\% of these sweeps, $D$ never decreases as the perturbation strength grows on either side of $s{=}1$.
Permitting decreases within each sweep's own noise floor raises this fraction to 99.0\%.
The per-sweep Spearman correlation $\rho$ between $d_{\boldsymbol{\Psi}}$ and $D$ has a median of 0.977.
In contrast, for the sweeps that violate the contraction condition, the fraction of sweeps in which $D$ never decreases drops to 49.0\%, and 10 of the 102 sampled sweeps produce empty renderings.
This degradation supports the contraction constraint of FineFractalDB as a condition under which local parameter distances remain structurally meaningful.
Taken together, these results establish $D$ as a reliable reference for structural difference and show that $d_{\boldsymbol{\Psi}}$ tracks it closely under local generative perturbations.

\subsection{Qualitative Examples of Filtered Triplets}
\label{sec:supp_triplet_examples}

\begin{figure*}[t]
  \centering
  \begin{subfigure}[t]{0.49\linewidth}
    \includegraphics[width=\linewidth]{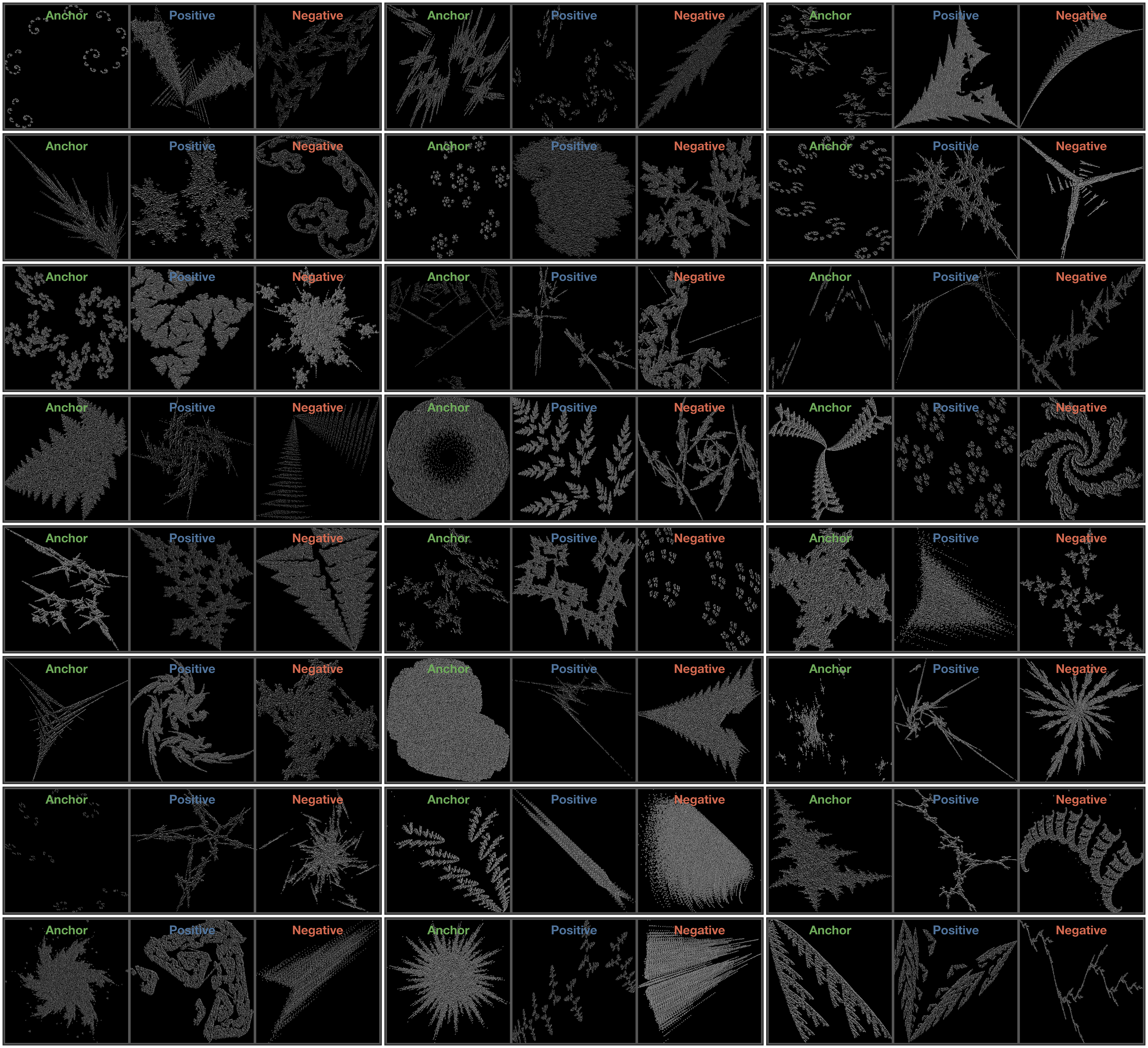}
    \caption{\textbf{all}}
  \end{subfigure}
  \hfill
  \begin{subfigure}[t]{0.49\linewidth}
    \includegraphics[width=\linewidth]{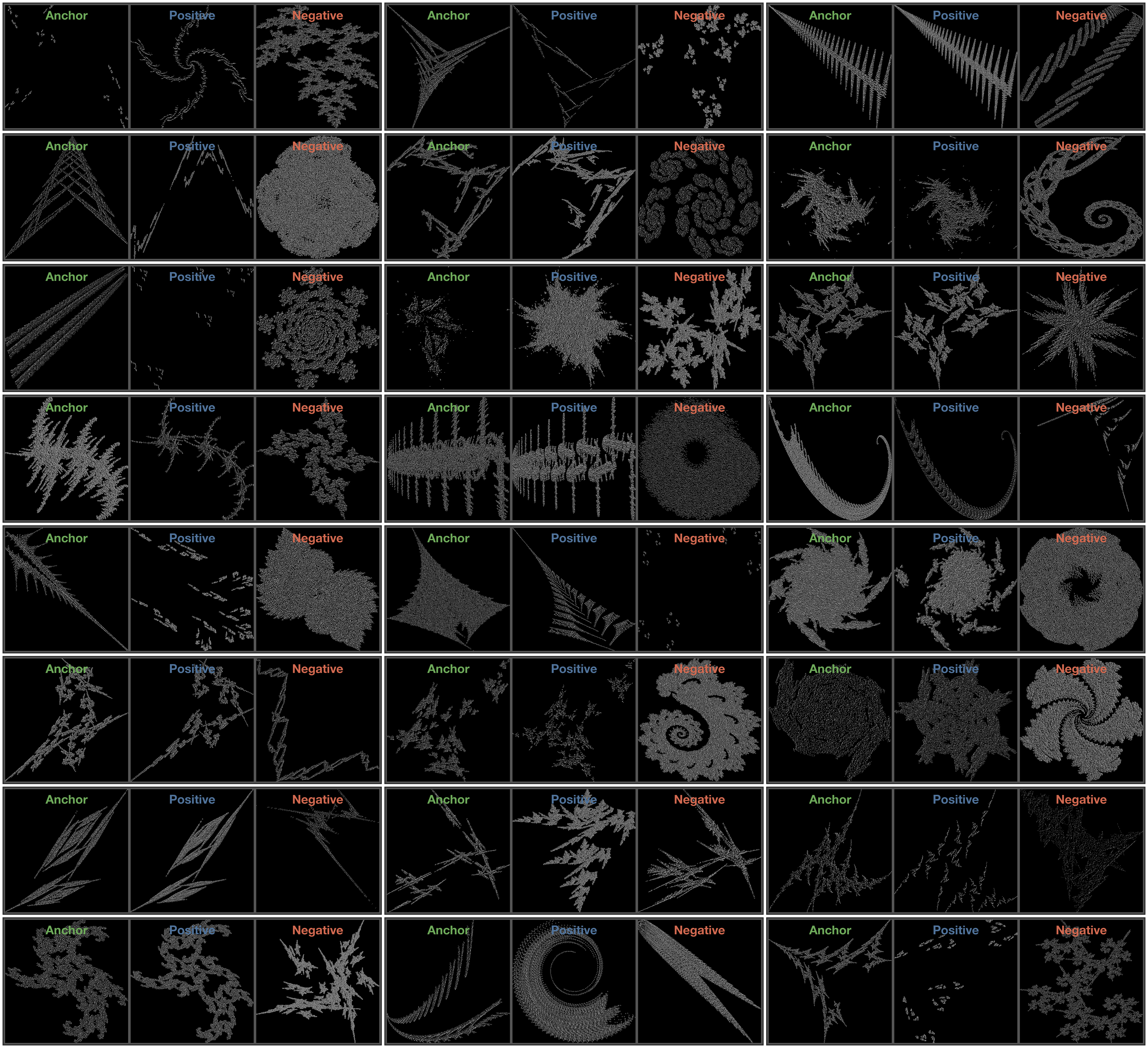}
    \caption{\textbf{each}}
  \end{subfigure}\\[2pt]
  \begin{subfigure}[t]{0.49\linewidth}
    \includegraphics[width=\linewidth]{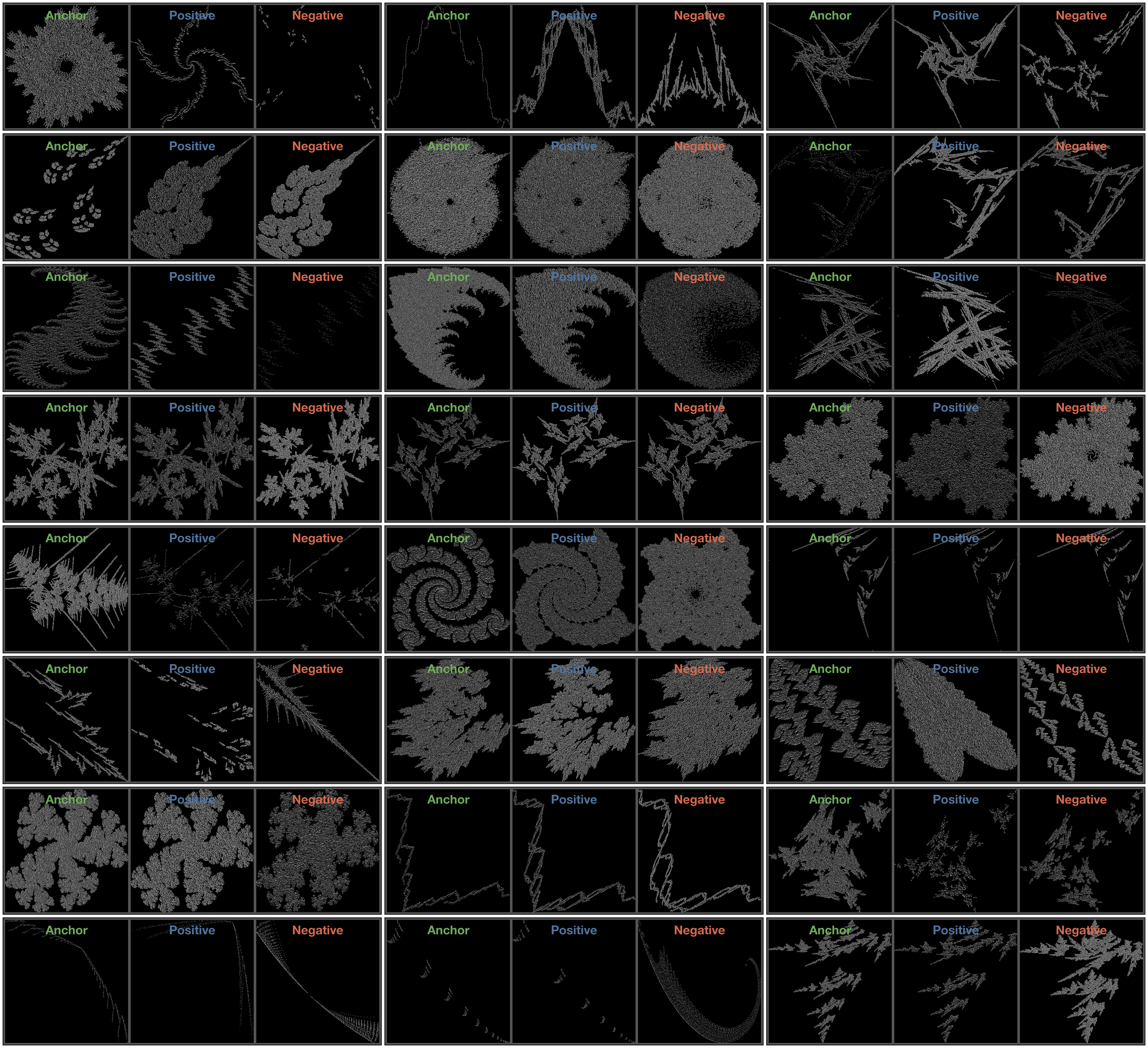}
    \caption{\textbf{same}}
  \end{subfigure}
  \hfill
  \begin{subfigure}[t]{0.49\linewidth}
    \includegraphics[width=\linewidth]{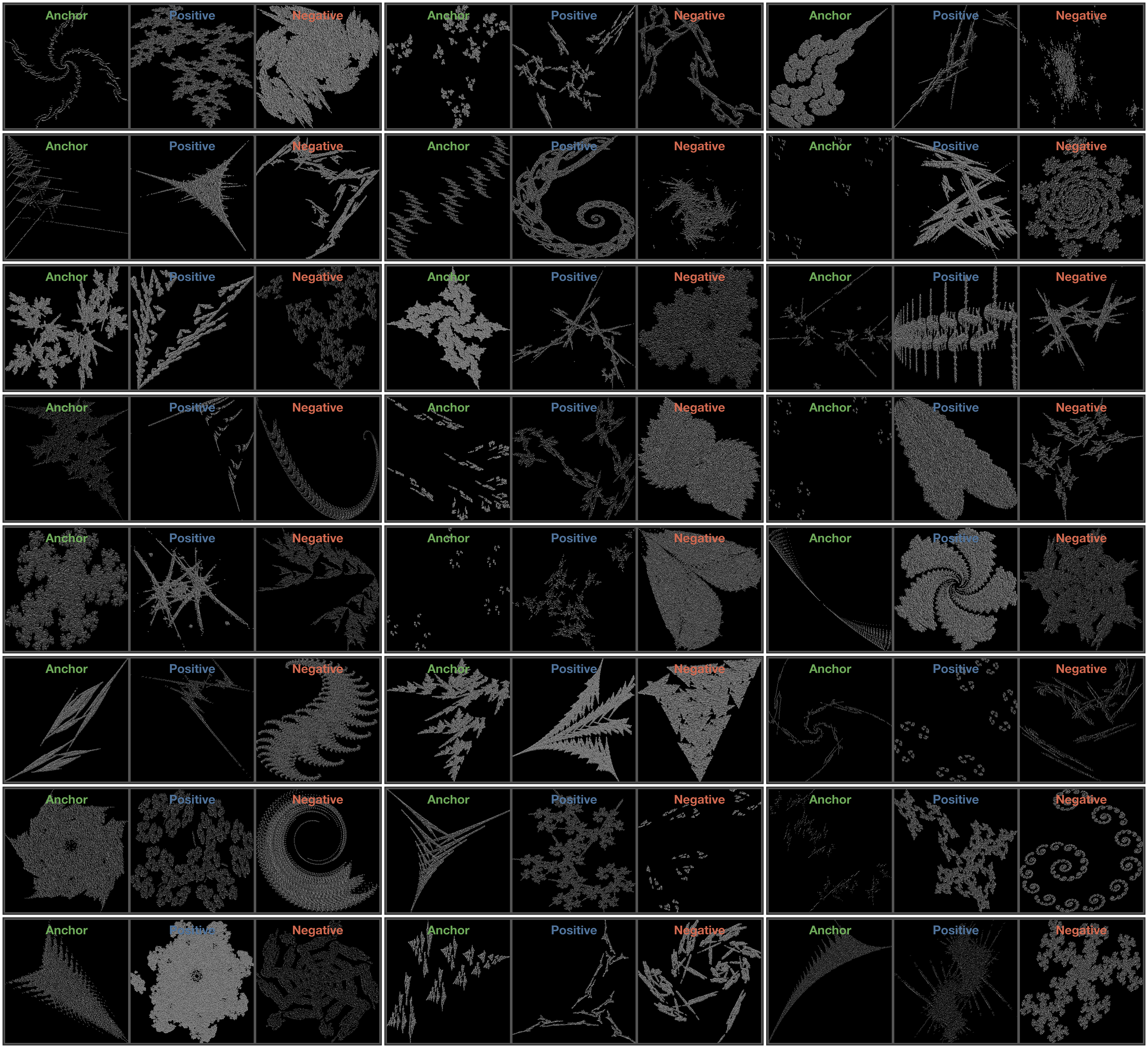}
    \caption{\textbf{diff}}
  \end{subfigure}
  \caption{Triplets sampled from FineFractalDB under the four filtering strategies (Sec.~3.4 of the main paper). Each cell shows an anchor and two candidates, with the positive and negative assigned by the descriptor distance $d_{\boldsymbol{\Psi}}$. \textbf{same} yields structurally close within-class candidates, \textbf{each} contrasts a within-class candidate with a cross-class one, and \textbf{all} and \textbf{diff} mostly produce candidates structurally unrelated to the anchor. Fine details are best observed when zoomed in.}
  \label{fig:supp_triplets}
\end{figure*}

\Cref{fig:supp_triplets} shows triplets sampled from FineFractalDB under the four filtering strategies (Sec.~3.4 of the main paper), allowing the descriptor-based assignment to be compared visually with the apparent structural similarity.
Under \textbf{same}, both candidates are drawn from the anchor's class and differ from it only in fine structural details, so ranking them is precisely the fine-grained decision that the descriptor is asked to make.
Under \textbf{each}, one candidate shares the anchor's class while the other does not, and the descriptor almost always selects the within-class candidate, consistent with its near-equivalence to the class-label rule reported in Sec.~5.1 of the main paper.
Under \textbf{all} and \textbf{diff}, the candidates are mostly unrelated to the anchor, and the ordering induced by $d_{\boldsymbol{\Psi}}$ no longer reflects structural similarity.
These examples make concrete why the reliability of the descriptor ordering depends on the candidate regime.

% ---------------------------------------------------------------------
% Figures

\section{FineFractalDB}

This section provides additional visual examples from FineFractalDB.
For comparison, we also present samples from the original FractalDB~\cite{Kataoka_2020_ACCV}.

\begin{figure*}[p]
    \begin{center}
        \centering
        \includegraphics[width=0.8\textwidth]{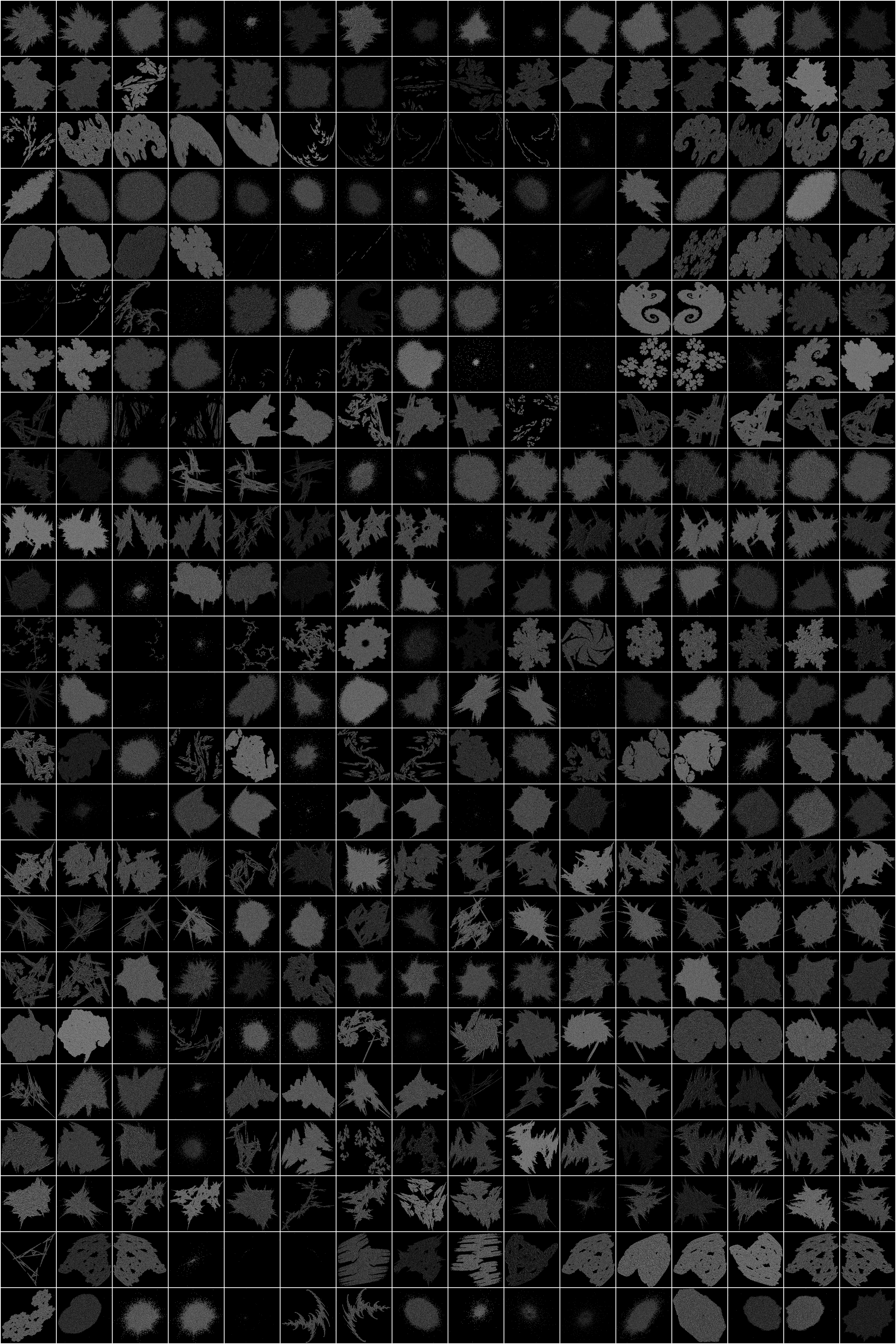}
        \caption{
            Randomly selected samples from FractalDB.
            Each row corresponds to a class and each column shows different instances belonging to that class.
            Fine details are best observed when zoomed in.
        }
        \label{fig:sup:randomly_picked_fractaldb}
    \end{center}
\end{figure*}

\Cref{fig:sup:randomly_picked_fractaldb} shows randomly selected images from FractalDB.
Because affine transformations are not strictly constrained to be contractive, many generated samples exhibit degraded fractal structures where characteristic patterns collapse into visually indistinct blobs.
In addition, some variations arise from discrete transformations such as flips and rendering patterns.

\begin{figure*}[p]
    \begin{center}
        \centering
        \includegraphics[width=0.8\textwidth]{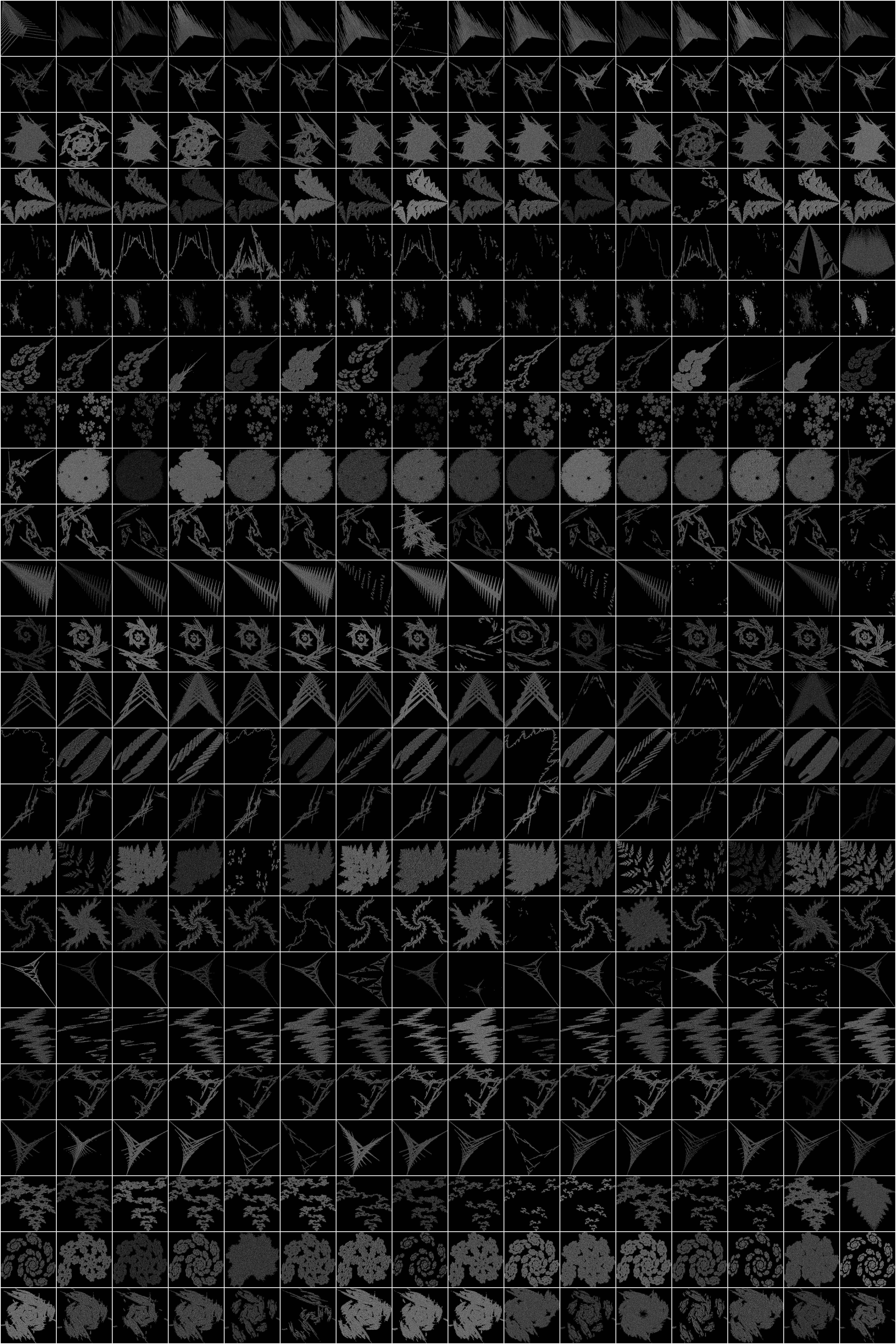}
        \caption{
            Randomly selected samples from the proposed FineFractalDB.
            Each row corresponds to a class and each column shows different instances generated from independently sampled parameters within the class.
            Fine details are best observed when zoomed in.
        }
        \label{fig:sup:randomly_picked_finefractaldb}
    \end{center}
\end{figure*}

\Cref{fig:sup:randomly_picked_finefractaldb} presents randomly selected images from the proposed FineFractalDB.
In contrast, FineFractalDB samples perturbations continuously from the parameter space while enforcing contraction constraints, resulting in clearer fractal structures.
Since the variations are generated through continuous perturbations, the visual differences between nearby instances can appear relatively small when only a limited number of samples are displayed.

\section{Qualitative Analysis}

\subsection{Grad-CAM Visualization}

\begin{figure*}[p]
    \begin{center}
        \centering
        \includegraphics[width=1.0\textwidth]{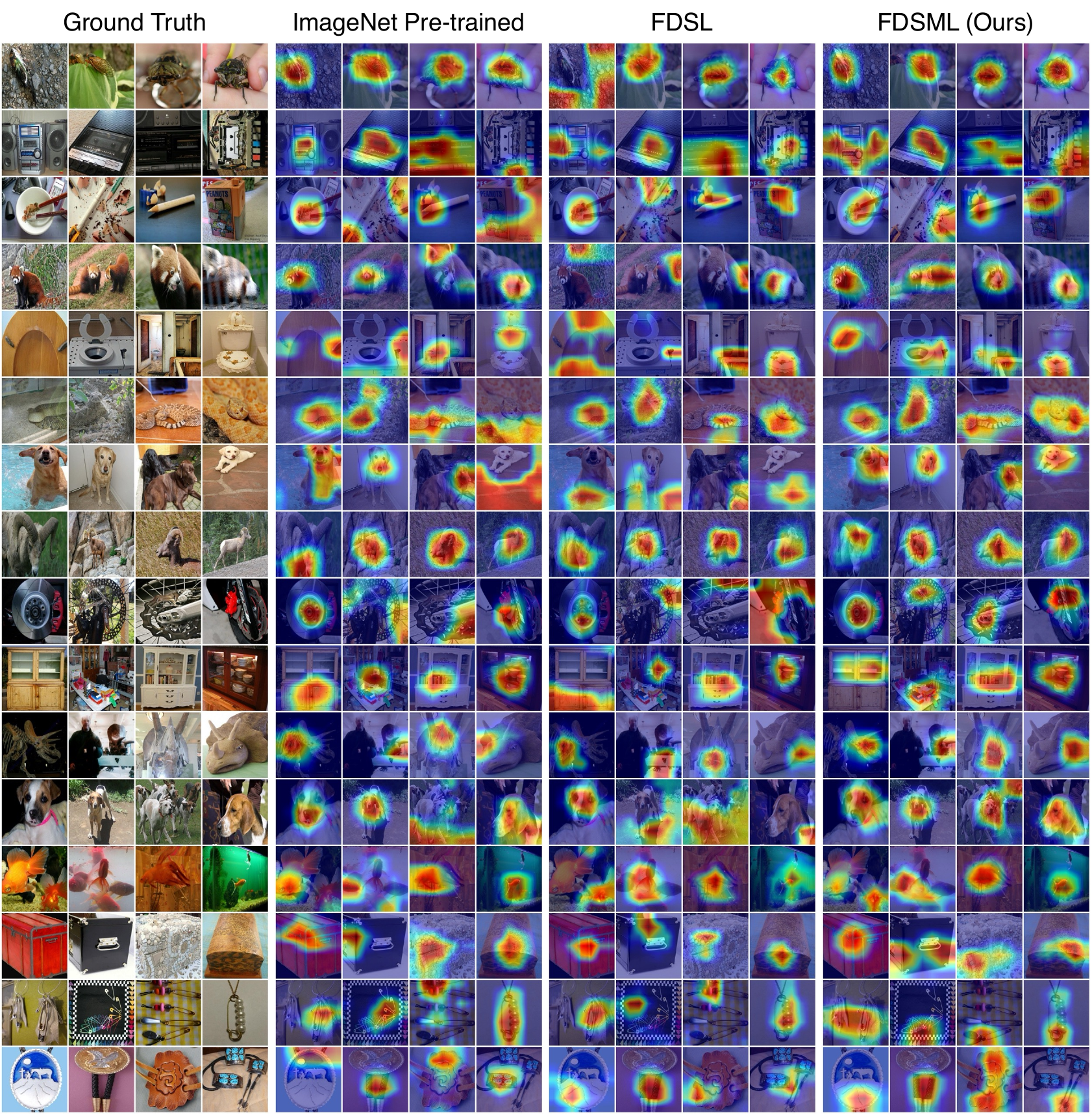}
        \caption{
            Grad-CAM visualizations for ResNet-50 models trained under different pre-training strategies.
            The leftmost panel shows the input images, while the subsequent panels present the corresponding Grad-CAM heatmaps produced by the ImageNet-1k pre-trained model, the FDSL-pre-trained model, and the proposed FDSML-pre-trained model.
            For the FDSL and FDSML settings, the models are first pre-trained on synthetic fractal datasets and subsequently fine-tuned on ImageNet-1k.
            Within each panel, rows correspond to object classes and columns correspond to different instances from the same class.
        }
        \label{fig:sup:grad_cam}
    \end{center}
\end{figure*}

\begin{figure*}[p]
    \begin{center}
        \centering
        \includegraphics[width=0.9\textwidth]{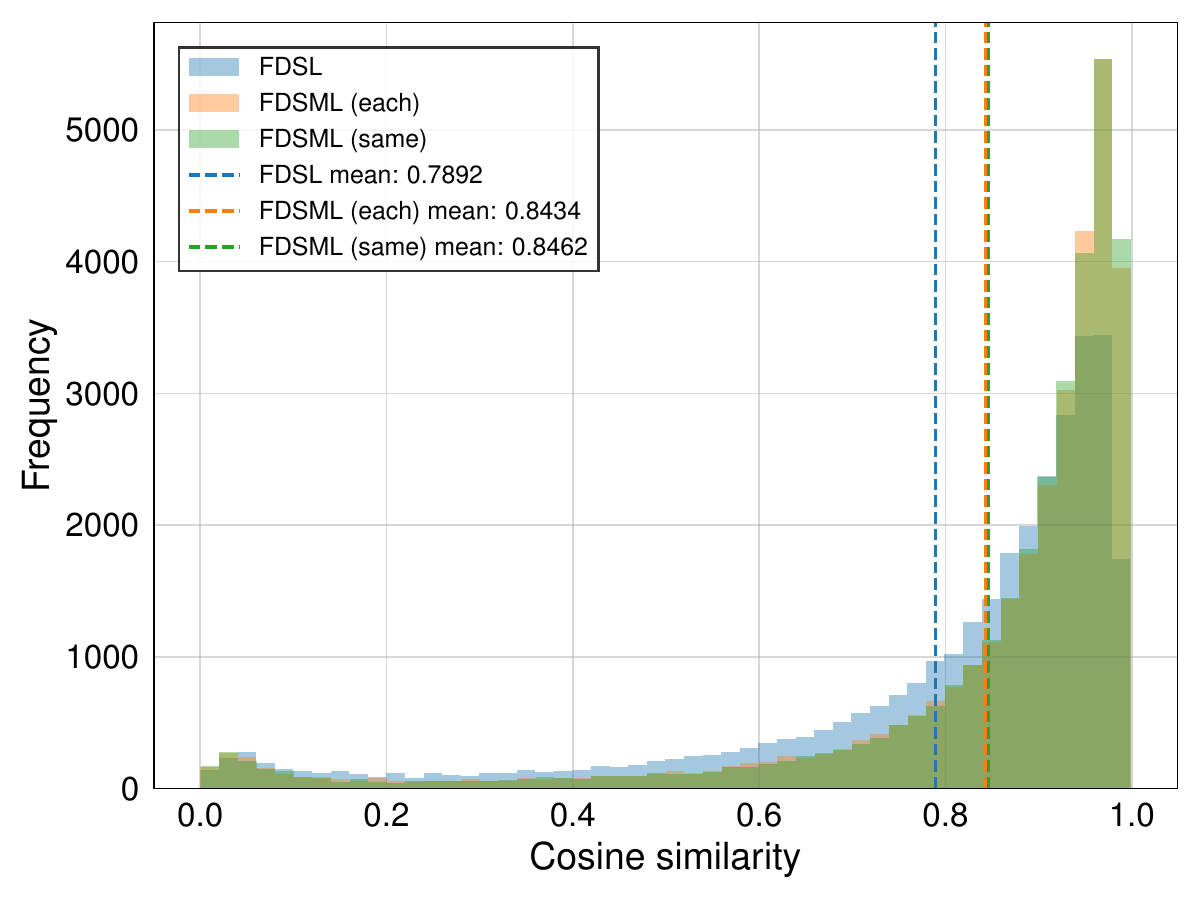}
        \caption{
            Histogram of cosine similarity between Grad-CAM maps produced by models pre-trained with different strategies and those from an ImageNet-pre-trained ResNet-50, used here as a reference model.
            Each similarity is computed per image using cosine similarity between Grad-CAM maps.
            Dashed vertical lines indicate the mean similarity for each method.
            FDSML variants exhibit higher similarity to the reference model than FDSL, suggesting that their activation patterns tend to align more closely with those of the natural-image pre-trained model.
        }
        \label{fig:sup:grad_cam_similarity}
    \end{center}
\end{figure*}

This section provides additional Grad-CAM visualizations for ResNet-50.
\Cref{fig:sup:grad_cam} presents representative examples obtained under different pre-training strategies.

When the ImageNet-pre-trained model is taken as a reference, models trained with synthetic data using FDSML exhibit activation patterns that more closely resemble those of the ImageNet-pre-trained model compared with FDSL.
This observation suggests that the proposed formulation encourages the model to capture object-relevant visual structures more effectively.

Interestingly, the ImageNet-pre-trained model occasionally produces activation maps that highlight background regions rather than the target object.
In many such cases, the FDSML-pre-trained model instead focuses on the object itself.
This behavior indicates that metric-learning-based supervision can encourage the model to rely more consistently on structurally meaningful cues for recognition.

To further quantify this observation, we compute the cosine similarity between the Grad-CAM maps produced by each model and those generated by the ImageNet-pre-trained model, which serves as a reference.
The resulting distributions are shown in \cref{fig:sup:grad_cam_similarity}.
As can be seen, the similarity scores for FDSML are consistently higher than those for FDSL, indicating that the activation patterns learned through the proposed formulation tend to align more closely with those of the natural-image pre-trained model.
In particular, the mean similarity increases from 0.7892 for FDSL to 0.8434 and 0.8462 for FDSML (each) and FDSML (same), respectively.
These results quantitatively support the qualitative observations in \cref{fig:sup:grad_cam}, suggesting that FDSML encourages representations that attend to semantically relevant regions of the image.

\subsection{Visualization of Attention Maps}

\begin{figure*}[p]
    \begin{center}
        \centering
        \includegraphics[width=0.7\textwidth]{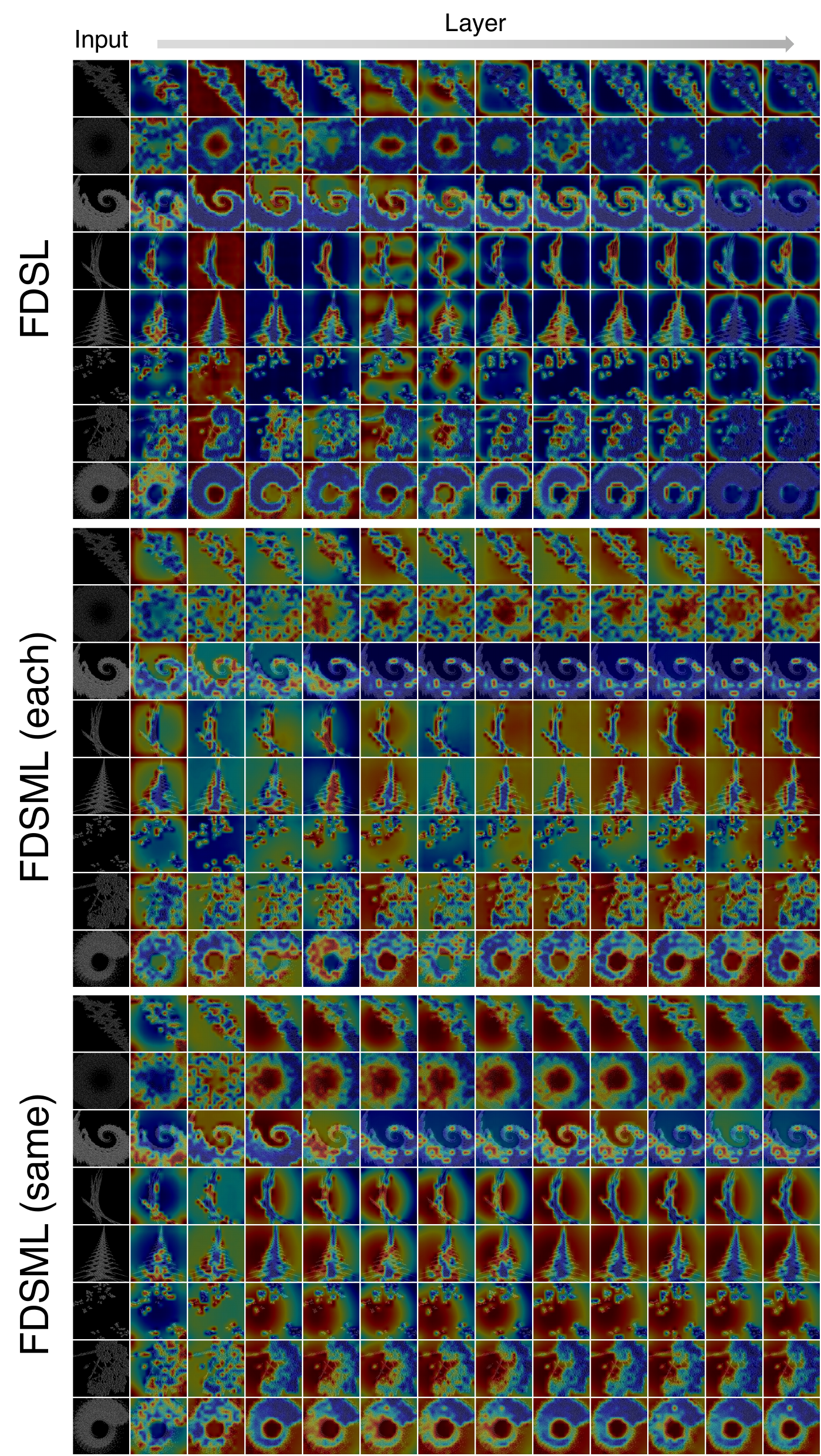}
        \caption{
            Layer-wise CLS-to-patch attention maps of Vision Transformer models after pre-training under different strategies.
            The leftmost column shows the input images, and the subsequent columns visualize attention maps from successive Transformer layers.
            Fine details are best observed when zoomed in.
        }
        \label{fig:sup:layerwise_attn}
    \end{center}
\end{figure*}

This section presents additional visualizations of layer-wise attention maps for Vision Transformer models after pre-training under different strategies.
\Cref{fig:sup:layerwise_attn} shows the CLS-to-patch attention across Transformer layers for the pre-trained models.

A clear difference can be observed between the two pre-training formulations.
The model pre-trained with FDSL tends to concentrate attention primarily along object boundaries.
In contrast, the model pre-trained with FDSML allocates attention not only to edges but also to textured regions within the object, indicating that the proposed formulation encourages the model to capture both structural and appearance cues.

Comparing the two variants of FDSML further reveals a difference in the types of visual cues emphasized during pre-training.
The model trained with FDSML (each) exhibits stronger attention on textured regions than FDSML (same).
A possible explanation lies in the scale of structural variations emphasized during training.
In the \textbf{same} strategy, candidate samples are drawn from the same class, requiring the model to distinguish subtle structural differences.
In contrast, the \textbf{each} strategy compares samples from different classes where larger structural variations are present.
As a result, FDSML (each) tends to rely more on appearance cues such as textures, whereas FDSML (same) focuses more on coarse structural boundaries.

\subsection{Visualization of Learned Embedding Space}

\subsubsection{Visualization of Embedding Space.}
\begin{figure*}[t]
    \begin{center}
        \centering
        \includegraphics[width=1.0\textwidth]{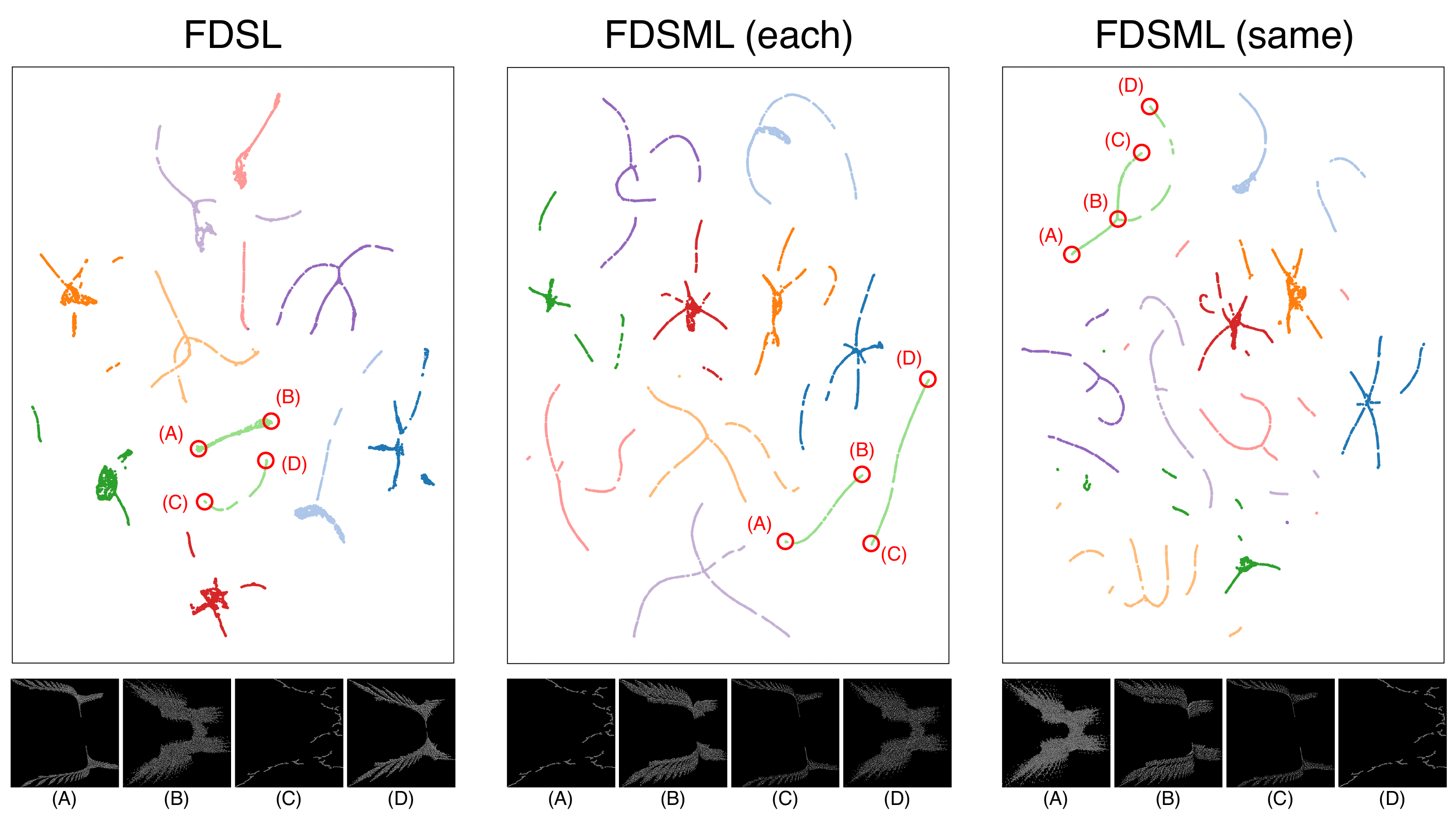}
        \caption{
            UMAP visualization of feature embeddings from ResNet-50 models pre-trained with different strategies.
            Features from the layer before the final fully connected layer are projected into two dimensions using UMAP~\cite{mcinnes2018umap-software}.
            Ten classes from the FineFractalDB test set are randomly selected and color-coded.
            Compared with FDSL, FDSML (each) forms more structured intra-class trajectories while remaining relatively well separated across classes, whereas FDSML (same) exhibits more noticeable overlap between classes.
            Points (A)–(D) indicate representative samples along a trajectory within the same class, with the corresponding fractal images shown below each panel.
        }
        \label{fig:sup:umap}
    \end{center}
\end{figure*}

To gain further insights into the structure of the learned representations, we visualize the embedding space using UMAP~\cite{mcinnes2018umap-software}, as shown in \cref{fig:sup:umap}.
For each pre-training strategy, features extracted from the layer before the final fully connected layer of the pre-trained ResNet-50 are projected into two dimensions.
We randomly select 10 classes from the FineFractalDB test set and visualize their embeddings with different colors.

We analyze the embedding structure in terms of inter-class separation and intra-class structure.
Compared with FDSL, the model pre-trained with FDSML (each) forms more structured trajectories within each class while remaining well separated across classes.
As highlighted by the samples (A)--(D), which correspond to representative points along a trajectory within a single class, the associated fractal images vary gradually along the trajectory, 
showing that gradual structural variation is also visible in the projected embedding.
The inter-class separation itself is expected in this regime, because the positive and negative assignments under each nearly coincide with the class-label rule, making the supervision effectively equivalent to class-label supervision.
The structured intra-class trajectories, in contrast, are not directly prescribed by such categorical supervision, suggesting 
qualitative differences in how intra-class variation is organized under the relational objective.

In contrast, the embeddings learned with FDSML (same) exhibit a greater degree of overlap across classes.
Since this strategy derives supervisory signals only from intra-class variations, the model may learn a more continuous representation of structural variations that extend beyond discrete class boundaries.

\subsubsection{Continuity of the Learned Embedding Space.}
To further examine whether the learned representations reflect the structure of the underlying generative process, we analyze the continuity of the embedding space induced by gradual changes in the fractal generation parameters.
Specifically, we randomly select two parameter sets from FineFractalDB and generate intermediate parameters by linear interpolation,
\begin{equation}
    \boldsymbol{\psi}(\alpha) = (1-\alpha)\boldsymbol{\psi}_{0} + \alpha\boldsymbol{\psi}_{1},
    \notag
\end{equation}
where the interpolation coefficient \(\alpha \in [0,1]\) is sampled at 5000 evenly spaced intervals.
Fractal images rendered from these parameters form a sequence that smoothly varies in the generative parameter space.

\begin{figure*}[t]
    \begin{center}
        \centering
        \includegraphics[width=0.8\textwidth]{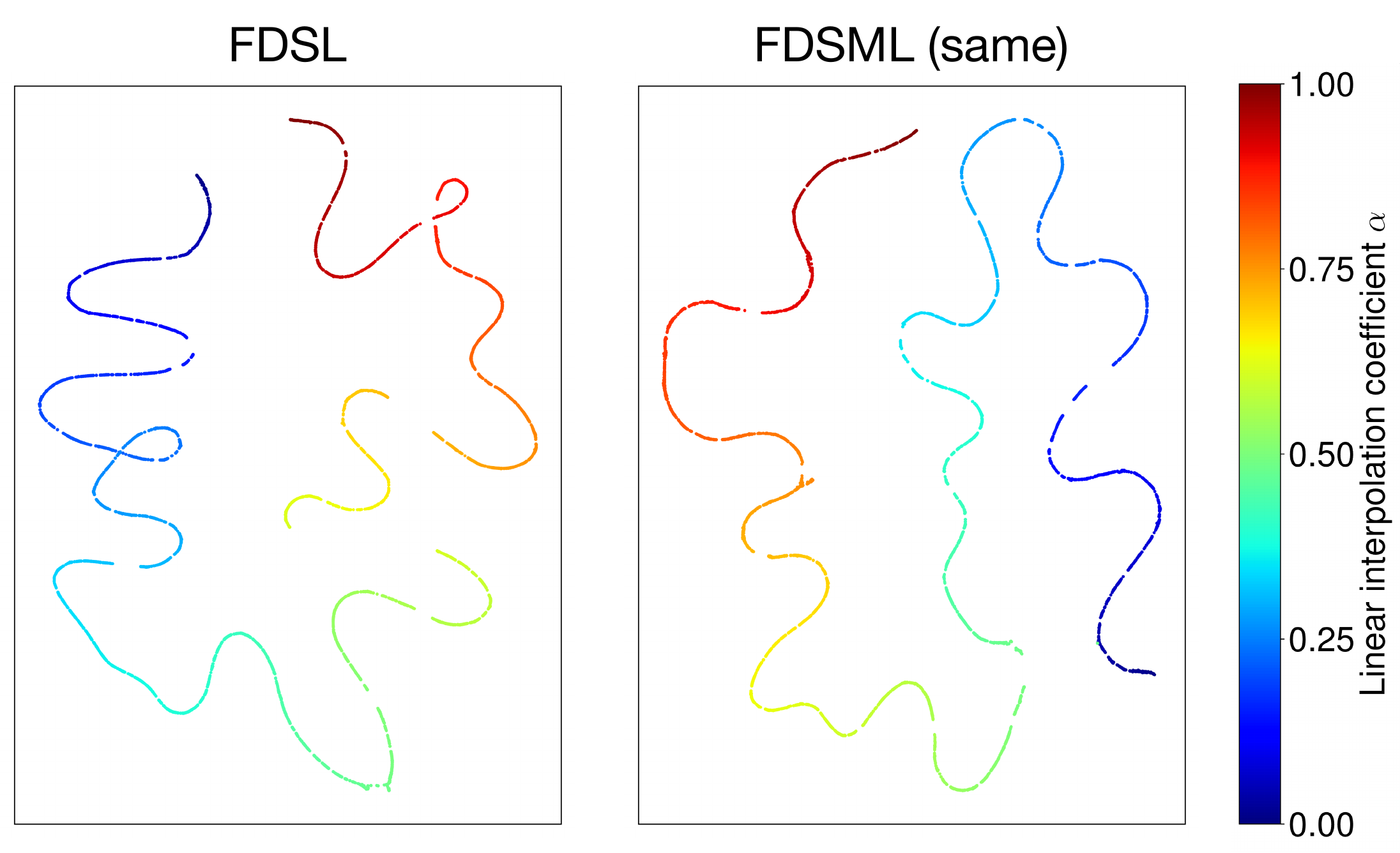}
        \caption{
            Visualization of embedding trajectories induced by linear interpolation in the generative parameter space.
            Two parameter sets are randomly selected from FineFractalDB, and intermediate parameters are generated by linear interpolation with coefficient $\alpha \in [0,1]$, sampled at 5000 evenly spaced intervals.
            Fractal images rendered from these parameters are then encoded using ResNet-50 models pre-trained with FDSL or FDSML (same), and the resulting embeddings are projected to two dimensions using UMAP.
            Points are colored according to the interpolation coefficient $\alpha$.
            While the projected embeddings of FDSL exhibit an abrupt trajectory transition around $\alpha \approx 0.6$, those of FDSML (same) form a smoother trajectory that more closely follows the gradual parameter interpolation.
        }
        \label{fig:sup:umap_lerp}
    \end{center}
\end{figure*}

Each generated image is encoded using ResNet-50 models pre-trained with either FDSL or FDSML (same).
The resulting embeddings are projected into two dimensions using UMAP~\cite{mcinnes2018umap-software}, and the embedding trajectories are visualized in \cref{fig:sup:umap_lerp}.
Points are colored according to the interpolation coefficient \(\alpha\), allowing the progression along the parameter interpolation to be observed in the embedding space.

As shown in \cref{fig:sup:umap_lerp}, the UMAP projection of the FDSL-pre-trained embeddings exhibits an abrupt transition around $\alpha \approx 0.6$, where the trajectory shifts markedly despite the smooth variation in the underlying generative parameters. In contrast, the FDSML (same) embeddings form a smooth trajectory that follows the progression of the interpolation parameter. Although this visualization alone does not establish continuity in the original high-dimensional embedding space, the projected trajectory under FDSML (same) is qualitatively more consistent with the gradual variation of the generative process than that under FDSL.

\section{Robustness Analysis}
\begin{table}[!tb]
  \caption{Classification Error (CE $\downarrow$, \%) on CIFAR-10-C with ResNet-50. Each entry reports the average CE over corruption types within each group.}
  \label{tab:cifar10-c-resnet}
  \centering

  \resizebox{1.0\linewidth}{!}{%
   \begin{tabular}{llccccc}
        \toprule
        PT Dataset & PT Type & Noise & Blur & Weather & Digital & Geometric \\
        \midrule
        FractalDB-1k & FDSL & \underline{39.69} & 35.49 & 27.07 & 22.15 & 21.55 \\
        \rowcolor{ldblue}
        FineFractalDB & FDSML (all)  & \textbf{38.24} & \underline{27.43} & \underline{22.73} & \textbf{19.37} & 18.84 \\
        \rowcolor{ldblue}
        FineFractalDB & FDSML (each) & 43.24 & \textbf{27.08} & 22.85 & 20.03 & \underline{18.73} \\
        \rowcolor{ldblue}
        FineFractalDB & FDSML (same) & 41.15 & 27.65 & \textbf{22.24} & \underline{19.69} & \textbf{18.47} \\
        \rowcolor{ldblue}
        FineFractalDB & FDSML (diff) & 42.55 & 28.66 & 24.06 & 20.27 & 19.26 \\
        \bottomrule
        \end{tabular}%
  }
\end{table}

\begin{table}[!tb]
  \caption{Classification Error (CE $\downarrow$, \%) on CIFAR-10-C with ViT-Tiny. Each entry reports the average CE over corruption types within each group.}
  \label{tab:cifar10-c-vit}
  \centering

  \resizebox{1.0\linewidth}{!}{%
        \begin{tabular}{llccccc}
        \hline
        PT Dataset & PT Type & Noise & Blur & Weather & Digital & Geometric \\
        \hline
        FractalDB-1k & FDSL         & 13.56 & 13.70 & 7.82 & 10.09 & 9.99\\
        \rowcolor{ldblue}
        FineFractalDB & FDSML (all)  & \underline{12.33} & \textbf{9.94} & \underline{6.24} & \underline{7.56} & \underline{7.78}\\
        \rowcolor{ldblue}
        FineFractalDB & FDSML (each) & \textbf{11.64} & 10.70 & 6.33 & \textbf{7.30} & 8.40 \\
        \rowcolor{ldblue}
        FineFractalDB & FDSML (same) & 12.45 & 10.31 & \textbf{6.15} & 7.58 & \textbf{7.77} \\
        \rowcolor{ldblue}
        FineFractalDB & FDSML (diff) & 13.07 & \underline{10.14} & 6.70 & 7.88 & 8.03  \\
        \hline
    \end{tabular}%
  }
\end{table}

We further evaluate the robustness of the learned representations against common visual corruptions using the CIFAR-10-C benchmark~\cite{hendrycks2019robustness}. 
Following the standard protocol, corruptions are grouped into five semantic categories: Noise, Blur, Weather, Digital, and Geometric. 
In CIFAR-10-C, these groups include Noise (Gaussian, Shot, Impulse, Speckle), Blur (Gaussian, Glass, Motion, Zoom), Weather (Snow, Frost, Fog, Spatter), Digital (Brightness, Contrast, Saturate, Pixelate, JPEG), and Geometric (Elastic). 
For reference, the ImageNet-C benchmark used in the main paper adopts a similar grouping with slightly different corruption types: Noise (Gaussian, Shot, Impulse), Blur (Defocus, Glass, Motion, Zoom), Weather (Snow, Frost, Fog), Digital (Brightness, Contrast, Pixelate, JPEG), and Geometric (Elastic). 
\Cref{tab:cifar10-c-resnet,tab:cifar10-c-vit} report the classification error (CE) averaged over corruption types within each group.

\subsection{Robustness Improvement on ResNet-50.}
As shown in \Cref{tab:cifar10-c-resnet}, FDSML reduces classification error relative to FDSL in the Blur, Weather, Digital, and Geometric groups for every variant.
For example, FDSML (each) reduces the Blur error from 35.49\% to 27.08\%, and FDSML (same) reduces the Geometric error from 21.55\% to 18.47\%.
These improvements may arise from the different training objectives of the two formulations.
While classification-based pre-training primarily encourages separability between discrete classes, FDSML learns to encode relative structural relationships between samples.
The learned representations may therefore rely less on fragile local cues and instead capture more stable structural cues that remain consistent under these corruptions.
The Noise group is the exception, as only FDSML (all) improves over FDSL (38.24\% versus 39.69\%) while the other variants perform worse.
We analyze this behavior in \cref{sec:supp_variants}.

\subsection{Robustness Across Architectures.}
The improvements extend to the ViT-Tiny architecture, as shown in \Cref{tab:cifar10-c-vit}.
On CIFAR-10-C, every FDSML variant reduces classification error relative to FDSL in all five corruption groups, and unlike ResNet-50, this includes the Noise group.
For example, FDSML (all) reduces the Blur error from 13.70\% to 9.94\%, and FDSML (each) reduces the Digital error from 10.09\% to 7.30\%.
Although Vision Transformers may inherently exhibit stronger robustness to certain corruptions, FDSML still provides consistent additional gains on this benchmark.
These results complement the ImageNet-C evaluation, where ViT-Tiny instead degrades in the Noise and Digital groups (Table~6 of the main paper).
Taken together, the two evaluations suggest that the robustness benefits of FDSML are not limited to convolutional backbones, although the direction of individual group-level effects depends on both the benchmark and the architecture.

\subsection{Comparison within Proposed Variants.}
\label{sec:supp_variants}
Among the proposed variants, FDSML (all) ranks first or second in nine of the ten combinations of architecture and corruption group, and FDSML (same) does so in four of the five groups on ResNet-50.
For ResNet-50, FDSML (same) achieves the lowest error in the Weather and Geometric groups, whereas for ViT-Tiny, FDSML (each) attains the lowest error in the Noise and Digital groups.

The variants separate most clearly under the Noise corruptions on ResNet-50, where FDSML (all) is the only variant that outperforms FDSL.
One possible explanation is that the unconstrained pool exposes the loss to comparisons spanning the widest range of structural differences, encouraging features that remain stable across instances rather than features attuned to fine structural distinctions.
The more constrained pools of \textbf{each} and \textbf{same} concentrate training on narrower structural differences, and the resulting representations may be more sensitive to unstructured high-frequency perturbations.
On ViT-Tiny, however, FDSML (each) attains the lowest Noise error despite its constrained candidate pool.

FDSML (diff) yields the smallest gains in most corruption groups.
By construction, its triplets compare the anchor exclusively with candidates from other classes.
For such cross-class comparisons, the descriptor ordering agrees with the structural ordering below chance, and the strategy simultaneously excludes the within-class comparisons for which the ordering is reliable.

\clearpage

{
    \small
    \bibliographystyle{ieeenat_fullname}
    \bibliography{supplementary_material}
}